\documentclass[sigconf,nonacm, final]{acmart}

\makeatletter
\renewcommand{\maketag@@@}[1]{\hbox{\m@th\normalsize\normalfont#1}}%
\makeatother
\usepackage{balance}
\usepackage{flushend}
\usepackage{color, soul}
\usepackage{pifont}
\usepackage[table]{xcolor}
\usepackage{multirow}
\usepackage{graphicx}
\usepackage{enumitem}
\usepackage{subcaption}
\usepackage{tabularx,array}
\newcolumntype{Y}{>{\centering\arraybackslash}X} 

\newcolumntype{J}{>{\justifying\arraybackslash}X} 
\newcolumntype{M}[1]{>{\centering\arraybackslash}m{#1}}
\usepackage[linesnumbered,ruled,vlined]{algorithm2e}
\usepackage{geometry}
\usepackage{booktabs,multirow,xcolor,ragged2e}
\usepackage[normalem]{ulem}

\usepackage{arydshln}

\usepackage{placeins}  

\AtBeginDocument{%
  }

\setcopyright{acmlicensed}
\copyrightyear{2018}
\acmYear{2018}
\acmDOI{XXXXXXX.XXXXXXX}
\acmConference[Conference acronym 'XX]{Make sure to enter the correct
  conference title from your rights confirmation email}{June 03--05,
  2018}{Woodstock, NY}

\acmISBN{978-1-4503-XXXX-X/2018/06}
\begin{document}

\title{Unveiling the Vulnerability of Graph-LLMs: An Interpretable Multi-Dimensional Adversarial Attack on TAGs}

\author{Bowen Fan}
\affiliation{%
  \institution{Beijing Institute of Technology}
  \country{China}
}
\email{fan1085165825@gmail.com}

\author{Zhilin Guo}
\affiliation{%
  \institution{Shandong University}
  \country{China}
}
\email{frank04180@outlook.com}

\author{Xunkai Li}
\affiliation{%
  \institution{Beijing Institute of Technology}
  \country{China}
}
\email{cs.xunkai.li@gmail.com}

\author{Yihan Zhou}
\affiliation{%
  \institution{Beijing Institute of Technology}
  \country{China}
}
\email{z2005416z@163.com}

\author{Bing Zhou}
\affiliation{%
  \institution{Shenzhen Institute of Technology}
  \country{China}
}
\email{bingzhou6@gmail.com}

\author{Zhenjun Li}
\affiliation{%
  \institution{Shenzhen Institute of Technology}
  \country{China}
}
\email{15323940@qq.com}

\author{Rong-Hua Li}
\affiliation{%
  \institution{Beijing Institute of Technology}
  \country{China}
}
\email{lironghuabit@126.com}

\author{Guoren Wang}
\affiliation{%
  \institution{Beijing Institute of Technology}
  \country{China}
}
\email{wanggrbit@gmail.com}

\renewcommand{\shortauthors}{Trovato et al.}

\begin{abstract}
    Graph Neural Networks (GNNs) have become a pivotal framework for modeling graph-structured data, enabling a wide range of applications from social network analysis to molecular chemistry. By integrating large language models (LLMs), text-attributed graphs (TAGs) enhance node representations with rich textual semantics, significantly boosting the expressive power of graph-based learning. However, this sophisticated synergy introduces critical vulnerabilities, as Graph-LLMs are susceptible to adversarial attacks on both their structural topology and textual attributes. Although specialized attack methods have been designed for each of these aspects, no work has yet unified them into a comprehensive approach. In this work, we propose the Interpretable Multi-Dimensional Graph Attack (IMDGA), a novel human-centric adversarial attack framework designed to orchestrate multi-level perturbations across both graph structure and textual features. IMDGA utilizes three tightly integrated modules to craft attacks that balance interpretability and impact, enabling a deeper understanding of Graph-LLM vulnerabilities. Through rigorous theoretical analysis and comprehensive empirical evaluations on diverse datasets and architectures, IMDGA demonstrates superior interpretability, attack effectiveness, stealthiness, and robustness compared to existing methods. By exposing critical weaknesses in TAG representation learning, this work uncovers a previously underexplored semantic dimension of vulnerability in Graph-LLMs, offering valuable insights for improving their resilience. Our code and resources are publicly available at \href{https://anonymous.4open.science/r/IMDGA-7289}{https://anonymous.4open.science/r/IMDGA-7289}.
\end{abstract}

\ccsdesc[500]{Computing methodologies~Semi-supervised learning settings}
\ccsdesc[500]{Computing methodologies~Neural networks}

\keywords{Text-attributed graphs; Graph-LLMs; Adversarial attacks}


\maketitle

\section{Introduction}

In the contemporary landscape of data science, graphs are indispensable data structures for representing intricate and interactive entities \cite{Xia_2021}. Within domains such as citation networks \cite{wang2020microsoft, hu2020ogb, sen2008collective} and social media platforms \cite{app10041327, ijcai2018p142,li2022distilling}, where nodes are characteristically imbued with copious semantic content, text-attributed graphs (TAGs) distinguish themselves from traditional graphs by offering a more semantically enriched structural paradigm \cite{chien2021node, chen2024exploring, he2023harnessing}.
Coinciding with this, Graph Neural Networks (GNNs) \cite{2016Convolutional_ChebNet,luo2024classic,sun2023adpa} have rapidly developed into a powerful tool for modeling TAGs, effectively capturing the intricate interactions and profound semantic connections within graphs. This ability to understand both the structure and semantics of TAGs has led to outstanding performance in various downstream tasks, including the biological networks \cite{NIPS2017_f5077839,10.1093/bioinformatics/bty294,qu2023app_gnn_bio2} and recommendation systems \cite{ cai2023app_gnn_rec3,LightGCN,su2024dcl}.

With the ascendance of large language models (LLMs), encoding models such as Sentence-BERT \cite{reimers-gurevych-2019-sentence} and RoBERTa \cite{liu2019roberta} have been adapted to the graph domain, giving rise to the new innovative paradigm known as Graph-LLMs \cite{jin2024large, qin2023disentangled, chen2023label, liu2023graphprompt}. This approach transcends shallow textual encoding methods (e.g., skip-gram \cite{mikolov2013distributed} and BoW \cite{harris1954distributional}), thereby endowing node features in TAGs with more profound semantic information. Concurrently, this non-decoupled paradigm dismantles the conventional separation of text feature processing and model architecture designing, significantly enhancing the capability of representation learning on TAGs \cite{zhu2024efficient, chen2024llaga}.

However, recent investigations have underscored the inherent vulnerability of GNNs to adversarial examples, which are meticulously crafted by introducing subtle perturbations to the original input data \cite{goodfellow2014explaining, kurakin2018adversarial, sun2022adversarial}. 
In the conventional GNN settings, attacks typically involve alterations to the graph structure or node features. Notably, Graph Modification Attacks (GMAs) \cite{dai2018adversarial, zugner_adversarial_2019_metaattck, sun2020adversarial, xu2019topology} and Graph Injection Attacks (GIAs) \cite{wang2020scalable,zou2021tdgia,zheng2021graph, chen2022understanding}, as emblematic instances, pose formidable challenges to the robustness of GNNs.
In the context of TAGs, the incorporation of additional textual information into Graph-LLMs introduces new security concerns, suggesting that attack methodologies from the field of Natural Language Processing (NLP) may have a significant impact on the representation learning of existing TAGs \cite{jin2020bert, wang2022semattack, li2020bert}.

\begin{figure}[t]
  \centering
  \includegraphics[width=1\linewidth,height=0.5 \linewidth]{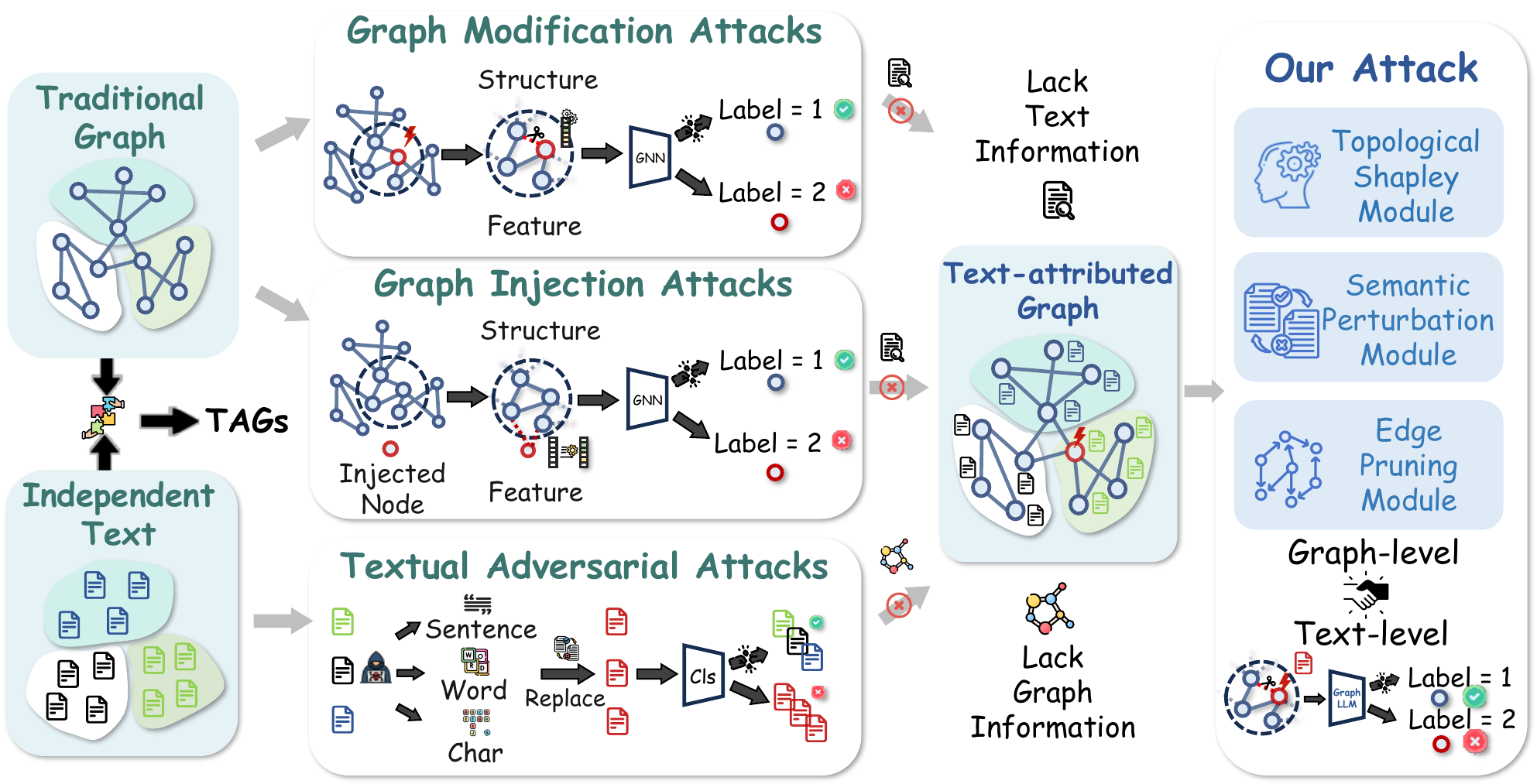}
  \caption{Illustration of different adversarial attack spaces}
  \vspace{-1em}
  \label{figure1}
\end{figure}

While prior adversarial attack methods have explored diverse perspectives (as illustrated in Figure \ref{figure1}), they clearly manifest the following shortcomings in scenarios integrating LLMs with graphs: \\
(1) \ul{\textit{ Incompatibility with real-world constraints:}} A large portion of conventional adversarial attacks rely on either unrestricted access to the target model's internal gradients or the deployment of poisoning data during the model training phase, both of which are often strictly unattainable in practical environments \cite{chen2020survey, chakraborty2018adversarial,sharma2018gradient}.\\
(2) \ul{\textit{ Limited scope in attack paradigms:}} Although GMAs and GIAs are meticulously designed attack frameworks that compromise GNNs by leveraging graph-structure knowledge, they fail to address vulnerabilities originating at the raw textual level \cite{lei2024intruding}. Conversely, while textual attacks from the NLP domain can significantly mislead models in classification, the textual features on TAGs are inherently interconnected through the graph's message-passing mechanism \cite{graphsurvey, xu2018gin}, making a naive direct transfer of these methods unviable. \\
(3) \ul{\textit{ Insufficient interpretability in graph attacks:}} Prevailing strategies primarily focus on perturbing node features and modifying the graph topology. However, these modifications often lack a human-centric perspective and suffer from limited interpretability, hindering a deeper understanding of Graph-LLMs' vulnerabilities and constraining the pursuit of more resilient defense approaches \cite{murdoch2019definitions, huang2020survey}.

\balance
To address the aforementioned limitations, we propose an \textbf{{I}}nter-pretable \textbf{{M}}ulti-\textbf{{D}}imensional \textbf{{G}}raph \textbf{{A}}ttack (IMDGA) framework in this study, with careful consideration of the algorithm's effectiveness, stealth, and interpretability.  
To ensure the algorithm aligns more closely with practical scenarios, we adopt a black-box setting \cite{papernot2017practical, mu2021hard}, wherein the attack leverages only limited information (e.g., model outputs) without any access to the internal parameters of the target model. Within the framework, IMDGA unfolds through three progressively connected stages to achieve effective attacks on TAGs. In the \textbf{warm-up stage}, IMDGA introduces the word-level \textit{Topological SHAP Module} to precisely quantify the contributions of salient tokens in graph information propagation \cite{ribeiro2016should, vstrumbelj2014explaining, lundberg2017unified}. This approach illuminates the semantic weight of each word from a human-centric perspective, allowing us to pinpoint pivotal words that exert varying degrees of influence on node classification predictions. 
Subsequently, during the \textbf{manipulation stage}, the framework incorporates the \textit{Semantic Perturbation Module}, which leverages the contextual understanding of the mask model to  generate a diverse pool of semantically plausible substitutes for the previously identified pivotal words \cite{garg2020bae}. A subsequent phase then employs a graph-aware scoring function to meticulously evaluate these candidates, quantifying their potential to disrupt predictions based on their topological and semantic relationships within the graph. This refined, two-stage process identifies and applies the optimal substitute, thereby introducing highly targeted perturbations while scrupulously preserving the attack's stealthiness.
To transcend the inherent limitations of existing graph text attacks and to achieve seamless integration with edge-level attack algorithms, we introduce a novel Influence-based \textit{Edge Pruning Module}  in the \textbf{refinement stage} to identify the most vulnerable edges, which are most susceptible to text-based perturbations \cite{muschalik2025exact, akkas2024gnnshap}. The strategic design of this module serves a dual function: it alleviates the computational bottleneck inherent in the Shapley strategy \cite{NIPS2017_7062} and refines the precision of the attack. By selectively targeting a minimal yet highly representative subset of samples, it significantly curtails computational overhead, while simultaneously identifying edges that exert a substantial and positive reinforcing influence on target node's classification confidence. 
This integrated approach, which strategically leverages the aforementioned three modules, not only advances the stealth and effectiveness of the attack but also provides an unprecedented level of interpretability, offering valuable insights into the most vulnerable components of Graph-LLMs.

The main contributions of our work are summarized as follows:
\ding{172} \ul{\textbf{\textit{A Novel Perspective on TAGs Security:}}} We present a pioneering and crucial perspective on the security vulnerabilities of TAGs, which arise from the non-decoupled nature of text encoding and GNN message passing. By synthesizing knowledge from both advanced NLP adversarial attacks and graph attack fields, we introduce a new paradigm for security analysis and attacks on TAGs.\\
\ding{173} \ul{\textbf{\textit{A Unified Multi-Dimensional Attack Framework:}}} We propose IMDGA, a holistic framework that effectively integrates three novel modules to perform interpretable attacks on both text and edge attributes. This methodology includes a human-centric approach for targeted text substitution and, critically, an edge-level attack strategy that transcends the inherent limitations of text-only perturbations, thereby further optimizing the attack's potency. \\
\ding{174} \ul{\textbf{\textit{Proven Effectiveness and Stealthiness:}}} The IMDGA method demonstrates exceptional attack success rates under stringent black-box conditions, showcasing robust performance across multiple datasets. The stealthiness of our attack is rigorously validated through empirical results and theoretical analysis, ensuring that the perturbations remain imperceptible while being highly effective.

\section{Background and Preliminary}
\captionsetup[figure]{}

\captionsetup[figure]{}
\begin{figure*}[t]
  \centering
  \includegraphics[width=0.85\linewidth,height=0.4\linewidth]{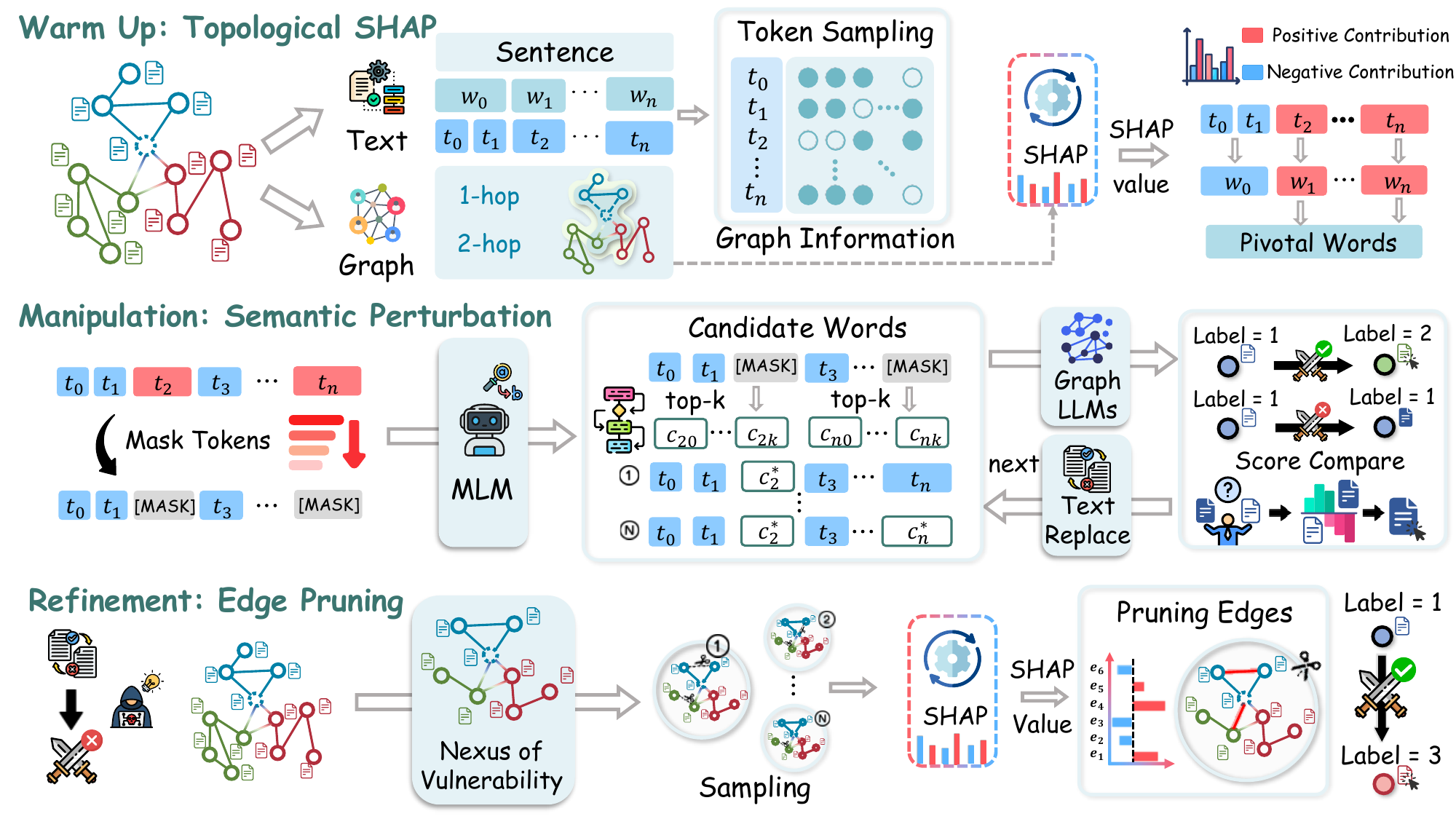}
  \caption{An overview of IMDGA framework, illustrating the key components and methodologies.}
  \vspace{-0.3cm}
  \label{Framework}
\end{figure*}

In this section, we will briefly introduce the key concepts and definitions to better explain the fundamentals of our proposed method.\\
\textbf{Text-Attributed Graphs.} Typically, a text-attributed graph is defined as $\mathcal{G} = (\mathcal{V}, \mathcal{E}, \mathcal{X}, \mathcal{T})$, where $\mathcal{V}$ denotes the set of nodes with $\vert \mathcal{V} \vert=n$ and $\mathcal{E} \subseteq \mathcal{V} \times \mathcal{V}$ represents the set of edges, which can also be represented by the adjacency matrix $\mathcal{A} \in \mathbb{R}^{n \times n}$. Any node $v_i \in \mathcal{V}$ has an associated text description $\mathcal{T}_i \in \mathcal{T}$. The feature matrix $\mathcal{X} \in\mathbb{R}^{n\times d}$, derived by encoding the texts $\mathcal{T}$, contains the feature vectors for all nodes, and $d$ represents the dimension of node feature.
Specifically, we define the label set as $\mathcal{Y} = \{y_1, y_2, \ldots, y_n\}$, where each label $y_i$ is uniquely associated with a node $v_i \in \mathcal{V}$.\\
\textbf{ LLM-based Graph Learning.} In this section, we elucidate the workflow of graph learning on TAGs when LLMs serve as enhancers. Broadly, such Graph-LLMs comprise three pivotal components: initialization, aggregation, and updating, which seamlessly integrate textual information with graph-structured data to enable efficient representation learning. For a given node $ v_i $ and its associated textual attribute $ t_i $, the transformer-based model generally encodes the raw text into a semantically rich embedding vector to initialize the node representation, expressed as follows:
\begin{equation}
h_{v_i} = x_i, \quad  x_i =  \psi_\theta(t_i) \in \mathbb{R}^d, \quad\forall v_i \in \mathcal{V},
\end{equation}
where $\psi_\theta$ denotes the text encoder with parameters $\theta$. To enrich the final representation of node $ v_i $, update operation $f_{Up}^{(k-1)}$ and aggregation function AGGR is employed to consolidate node representations derived from the preceding iteration: 
\begin{equation}
h_{vi}^{(k)} =f_{Up}^{(k-1)} (h_{v_i}^{(k-1)},\text{AGGR}(\{h_{u}^{(k-1)},u\in \mathcal{N}(v_i)\})).
\end{equation}\\
Embedded from raw text and then iteratively refined, these node representations capture both rich textual semantics and graph structure information, making them suitable for various downstream tasks such as node classification and link prediction.\\
\textbf{Adversarial Attack.}
The vulnerability of TAGs stems from two principal attack vectors: graph and text adversarial attacks. \\
Graph adversarial attacks are designed to deceive a GNN $\mathcal{F}_{\theta}$ by subtly altering the original graph $\mathcal{G}$, causing the model to produce incorrect predictions. This objective can be formally expressed as:
\begin{equation}
\max \mathcal{L}(\mathcal{F}_{\theta}(\mathcal{G'})), \quad \text{s.t.} \ \left\|\mathcal{G'} - \mathcal{G}\right\| \leq \Delta,
\end{equation}
where $ \mathcal{G'} $ denotes the perturbed graph and $\Delta$ denotes the perturbation budgets, containing $\Delta_A$ and $\Delta_X$.
In GMAs, modifications are restricted to making subtle perturbations on the existing adjacency matrix $\mathcal{A}$ and feature matrix $\mathcal{X}$. Specifically, these perturbations must satisfy the constraint $\left\|\mathcal{A}' - \mathcal{A}\right\|_{0} \leq \Delta_A$ and $\left\|\mathcal{X}' - \mathcal{X}\right\|_{\infty} \leq \Delta_X$. For GIAs, the graph structure is expanded by introducing malicious nodes $\mathcal{V}_{\text{atk}}$ with corresponding adversarial features $\mathcal{X}_{\text{atk}}$, resulting in $\mathcal{X}' = \begin{pmatrix} \mathcal{X} \\ \mathcal{X}_{\text{atk}} \end{pmatrix}$ and $\mathcal{A}' = \begin{pmatrix} \mathcal{A} & \mathcal{A}_{\text{atk}} \\ \mathcal{A}_{\text{atk}}^T & O_{\text{atk}} \end{pmatrix}$. Similarly, the number of attacked nodes $|\mathcal{V}_{atk}|$, the degree of nodes $d_v$, and the feature matrix $\mathcal{X}_{atk}$ are bounded to ensure stealthiness. \\
Text adversarial attacks are commonly employed in tasks such as text classification, aiming to craft adversarial samples for the textual attributes $ \mathcal{T}^\ast $ that significantly diminish the confidence of classifier $\Phi$ in accurate predictions, thereby inducing misclassification:
\begin{equation}
\mathcal{T}^*=\arg\min_{\mathcal{T}^{\prime}}\left\{-D(\mathcal{T},\mathcal{T}^{\prime})\right\}, \quad\mathrm{s.t.}\quad \Phi(\mathcal{T}^{\prime})\neq y,
\end{equation}
where D denotes the semantic similarity.
Drawing inspiration from the above adversarial attacks, we propose a novel perspective to combine their advantages, investigating the vulnerabilities of TAGs.\\
\textbf{Shapley Value.} SHAP (SHapley Additive exPlanation) is a game-theoretic framework that employs the concept of Shapley values \cite{shapley1953value, vstrumbelj2014explaining} to quantify the contribution of individual players (e.g., feature) in cooperative scenarios. Intuitively, the more pivotal a player is to the prediction, the higher its corresponding Shapley value will be. In general, the Shapley value of a player is computed as the weighted average of all possible marginal contributions that the player provides across different coalitions $S$:
\begin{equation}
    \phi(i)=\sum_{S\subseteq\{1,...,n\}\setminus\{i\}}^{2^{n-1}}\frac{|S|!(n-|S|-1)!}{n!}\left[f(S\cup\{i\})-f(S)\right],
    \label{shap}
\end{equation}
where $n$ denotes the total number of players and $f(S\cup{i})-f(S)$ quantifies the marginal contribution of player $i$ to the coalition $S$. From an explainable perspective, the raw text, features, or edges within TAGs can be seen as collaborative 'players' that jointly explain model predictions. This leads to the concept of Topological SHAP, which makes the attack process more interpretable.

\section{Methodology}

\balance
In this section, we present a comprehensive overview of our innovative method, which systematically probes the vulnerabilities of Graph-LLMs with a focus on interpretability. IMDGA introduces a pioneering paradigm for adversarial attacks on representation learning in TAGs under black-box conditions, achieving precise and efficient targeted manipulations. It strategically accounts for both textual attributes and graph structural dynamics in its attack design, addressing three important challenges through synergistic modules. Specifically, IMDGA first introduces the Topological SHAP to tackle \textbf{Challenge \ding{182}}: How to pinpoint pivotal words in the raw text that critically influence model predictions from a graph-centric perspective (Section 3.1)? Subsequently, it employs the Semantic Perturbation to address \textbf{Challenge \ding{183}}: How to execute stealthy, semantically coherent perturbations to these pivotal words (Section 3.2)? Finally, IMDGA proposes the Edge Pruning to achieve more advanced adversarial effects, resolving \textbf{Challenge \ding{184}}: How to precisely disrupt key message-passing pathways in a human-intuitive manner (Section 3.3)? For clarity, we illustrate the overall framework in Figure \ref{Framework} and provide pseudocode in Appendix \ref{app:code}.

\subsection{Topological SHAP (Warm Up)}
Words serve as the fundamental building blocks of sentences, embodying independent semantic units that often encapsulate core intent, such as subjects or sentiments. While numerous NLP techniques have pursued word-level textual attacks, the texts in TAGs differ markedly from those in prior studies as they are intrinsically tied to nodes, interconnected via edges, rendering them interdependent rather than isolated passages. Consequently, attacking a single node's text in isolation without accounting for its connected neighbors fails to fully exploit the graph's textual attributes. To address this, we introduce the Topological SHAP Module in the warm-up stage, which maps the SHAP framework onto graph structures to quantify word importance from a topological viewpoint: 
\begin{equation}
    \phi(i) = \!\!\!\!\sum_{S \subseteq \mathcal{W} \setminus \{W_i\}} \!\!\!\frac{|S|! (|\mathcal{W}| \!- |S| - 1)!}{|\mathcal{W}|!} \left[ f(\mathcal{T}_S) - f(\mathcal{T}_{S \cup \{W_i\}}) \right],
\end{equation}
where for a node $v$, $\mathcal{W} = \{W_i \mid W_i \in Tokenize(\mathcal{T}_v), i = 1, \dots, m\}$ denotes the subset of words derived from the node’s text $\mathcal{T}_v$, with $W_i$ representing the $i$-th word. The masked text $\mathcal{T}_S$ is defined as:
\begin{equation}
    \mathcal{T}_S = \left\{ W_i \cdot \mathbb{{I}}_{\{W_i \notin S\}} + [\text{MASK}] \cdot \mathbb{{I}}_{\{W_i \in S\}} \mid W_i \in \mathcal{W} \right\}.
\end{equation}
Notably, while $\mathcal{T}_S$ is defined as a set, it retains the original text sequence order when used as input.
In addition, the coalitional function $f(\mathcal{T}_S)$ is defined as the aggregation of the GNN’s predictive scores over $v$ and its neighborhood $\mathcal{N}(v)$, which serves as the basis for computing the marginal contribution:
\begin{equation}
    f(\mathcal{T}_S) = \sum_{u \in \{v\} \cup \mathcal{N}(v)} \mathcal{F}_\theta(\mathcal{G}, \psi_\theta(\mathcal{T}_S), u) , \quad \forall v \in \mathcal{V}.
\end{equation}
In contrast to traditional SHAP, our approach employs a mirrored operation, bypassing the conventional paradigm of incrementally adding features. By masking words, we reformulate the marginal contribution from the additive transition ($S $ to $S \cup \{i\}$ ) to a subtractive shift ($S \cup \{W_i\}$ to $S) $. The Shapley value $\phi(i)$ quantifies a word’s impact on the classification of both the node and its neighborhood, providing a precise measure of its significance in Graph-LLM predictions. Recognizing that $\phi(i)$ encompasses contributions across all classes, we refine the final score to focus on the sum of its Shapley values for the true labels of relevant nodes $\phi^{y_u}(i)$:
\begin{equation}
    \xi(i) = \sum_{u \in \{v\} \cup \mathcal{N}(v)} \phi^{y_u}(i), \quad \forall v \in \mathcal{V}.
\end{equation}
Once the importance score $\xi(i)$ of each word within a sentence has been obtained, we further refine the selection to identify the most influential tokens. Accordingly, we define the pivotal word set $\mathcal{P}$ as the top-$k$ words whose $\xi(i)$ values exceed a threshold $\tau$:
\begin{equation}
    \mathcal{P} = \{ W_i \mid \xi(i) > \tau, i \in \mathcal{I}_k \}.
\end{equation}
The pivotal word set $\mathcal{P}$ thus encapsulates the most semantically and topologically critical words that govern the prediction behavior of both the target node and its neighbors. By distilling the text into this compact subset of decisive words, we establish a principled foundation for manipulation stage, where carefully designed perturbations can be applied in a targeted and stealthy manner.

\balance
\subsection{Semantic Perturbation (Manipulation)}
Inspired by the limitations of conventional textual attacks, which fail to leverage inter-node message passing to amplify their impact through graph structures, we introduce the adversarial Semantic Substitution Module. This module generates a diverse pool of semantically plausible substitutes for the pivotal words in $\mathcal{P}$, meticulously balancing maximal disruption of model predictions with minimal surface-level detectability. In addition, it ingeniously transforms a Masked Language Model (MLM) into a context-aware "semantic proxy," transcending the limitations of traditional static substitutions. Crucially, since each encoder corresponds to its own specialized MLM, the resulting candidate words are highly aligned with the model's internal semantic representation. Drawing from MLM's pre-training objective, which maximizes the product of conditional probabilities for masked tokens given surrounding context:
\begin{equation}
\prod_{i=1}^{m} P(W_i \mid W_1, \dots, W_{i-1}, W_{i+1}, \dots, W_m).
\end{equation}
\\
Building upon this principle, our framework exploits such contextual dependency modeling to synthesize semantically coherent and contextually plausible candidate substitutions. Taking node $v$ as an example, we define its candidate word set $\mathcal{C}$ as the union of top-$k$ replacements for each pivotal word $W_i \in \mathcal{P}$,  where each pivotal word generates multiple semantically proximate candidates to ensure diversity and contextual adaptability. For each candidate $r \in \mathcal{C}$, we generate the perturbed text $\mathcal{T}' = (\mathcal{T} \setminus \{W_i\}) \cup \{r\}$. Subsequently, we compute GNN’s predictive probability distribution $p_u(\mathcal{G}, \mathcal{T}')$ for nodes $u \in \{v\} \cup \mathcal{N}(v)$. The confidence gap is defined as:
\begin{equation}
    \delta_u(r) = p_u^{(1)}(\mathcal{G}, \mathcal{T}') - p_u^{(2)}(\mathcal{G}, \mathcal{T}'),
    \label{gap}
\end{equation}
where $p_u^{(1)}$ and $p_u^{(2)}$ denote the largest and the second-largest predicted probabilities, respectively, quantifying the model’s decision certainty for node $u$ post-replacement. Evidently, a smaller confidence gap indicates a more ambiguous decision boundary for the model, thereby rendering it more vulnerable to adversarial attacks. Therefore, aggregating these gaps across the neighborhood yields the base score for candidate word $r$:
\begin{equation}
    \Delta(r) = \sum_{u \in \{v\} \cup \mathcal{N}(v)} \delta_u(r).
\end{equation}
To incorporate adaptive characteristics, we introduce the label-flip indicator function $\mathbb{{I}}_{\text{flip}}(r)$, which is 1 if the replacement induces a prediction label flip for node $v$ or its neighborhood, and 0 otherwise. The final replacement score is defined as:
\begin{equation}
    \sigma(r) = \Delta(r) \cdot \left(1 + \alpha \cdot \mathbb{{I}}_{\text{flip}}(r)\right),
\end{equation}
where $\alpha > 0$ is a hyperparameter that dynamically amplifies the confidence gap weight in label-flip scenarios, prioritizing high-impact replacements. The aggregated score $\Delta(r)$ ensures a comprehensive neighborhood evaluation, while the adaptive factor $\left(1 + \alpha \cdot \mathbb{{I}}_{\text{flip}}(r)\right)$ elegantly highlights boundary perturbations, balancing stability and sensitivity analysis. This innovative mechanism implicitly reinforces the graph interactions through neighborhood aggregation, endows the attack with greater strategic depth and robustness.\\

\subsection{Edge Pruning (Refinement)}
Although textual attacks effectively disrupt node representations by perturbing high-contribution words in node texts, their efficacy is limited in scenarios where graph topology strongly dominates GNN predictions. To address this, we propose an interpretable Edge Pruning Module as a strategic extension of textual perturbations, activated only when textual disruptions fail to induce target label flips. The core of this module lies in our innovative concept, nexus of vulnerability, which represents a curated subset of nodes highly intertwined with the target node $v$ and inherently susceptible to attacks. We identify this nexus through a robust multifaceted scoring mechanism that elegantly fuses three complementary dimensions: predictive disparity, feature influence, and vertex centrality. Drawing from the confidence gap $\delta_u(r)$ in Eq.~\eqref{gap}, we similarly define predictive disparity $\delta(u)$ for node $u$, quantifying decision boundary ambiguity. Critically, feature influence is derived from message propagation dynamics, measured via the L1 norm of the expected Jacobian for node $v$'s impact on $u$ after $k$ layers:
\begin{equation}
    I(u, v, k) = \left\| \mathbb{E} \left[ (\partial \mathcal{X}_u^{(k)}) / (\partial \mathcal{X}_v^{(0)}) \right] \right\|_1,
\end{equation}
where $X_u^{(k)}$ denotes node $u$'s feature after the $k$-th layer, and $X_v^{(0)}$ is node $v$'s initial feature embedding. Normalized, it yields:
\begin{equation}
I_u(v, k) = \frac{I(u, v, k)}{\sum_{w \in V} I(u, w, k)},
\end{equation}
reflecting node $v$'s relative contribution to node $u$'s representation. This allows us to quantify the feature influence from target node $v$ to other nodes. Typically, nodes with lower degrees, owing to their reduced structural redundancy, are more prone to amplified perturbation effects. To capture this property, we incorporate $\tfrac{1}{\deg(u)}$ as a surrogate indicator of structural centrality and attack susceptibility, thereby forming the final term in the computation of the vulnerability score.
We consolidate the above dimensions into a unified vulnerability scoring function, formally defined as:
\begin{equation}
\text{Score}(u) = \alpha_1 \cdot (1 - \delta(u)) + \alpha_2 \cdot I_u + \alpha_3 \cdot \left( \frac{1}{\deg(u)} \right),
\end{equation}
 Nodes with the highest scores form the nexus of vulnerability $\mathcal{G}_n(v)$, constraining the attack domain to a high-impact subgraph. On this basis, we delineate critical paths linking nodes in $\mathcal{G}_n(v)$ to $v$, and employ an efficient Shapley value approximation inspired by GNNShap, cast as a least-squares solution:
\begin{equation}\hat{\phi} = (M^\top U M)^{-1} M^\top U \hat{y},\end{equation}
where $M \in \mathbb{R}^{k \times n}$ is the mask matrix encoding $k$ subgraph samples across $n$ edges, $U$ is a diagonal weight matrix reflecting sample importance and $\hat{y}$ approximates nexus predictions under masked configurations. Finally, we prune the top-$k$ edges with the highest attributions, further eroding model confidence in node's original label. This interpretable pruning mechanism not only complements textual attacks but also uncovers topological vulnerabilities, forging a multi-dimensional adversarial strategy that advances the frontier of explainable TAGs attacks.

\subsection{Time Complexity Analysis}

To substantiate the scalability of IMDGA, we provide a theoretical analysis highlighting how interpretability is achieved without sacrificing computational efficiency. First, the exponential cost of exact Shapley computation is alleviated through partition-based sampling: for a node with $m$ tokens, the coalition space is reduced to $s \ll 2^m$. Denoting by $t_g$ the inference time of the underlying Graph-LLM, the complexity of the Topological SHAP Module becomes $O(s \cdot t_g)$. 
For token substitution, the Semantic Perturbation bounds computation by restricting to $ |\mathcal{P}|$ pivotal tokens, each producing $k_c$ semantically coherent candidates; since the masked language model requires only one forward pass, the cost is $O(k_c|\mathcal{P}| \cdot t_g + t_m)$, where $t_m$ is the MLM inference time. 
Finally, the Edge Pruning Module narrows the sampling space to a subgraph $\mathcal{G}_n(v)$ around the target node and further samples $k\ll 2^{|\mathcal{N}(v)|}$ coalitions within this restricted domain, which not only limits the computational scope but also improves the precision of Shapley-based attribution, yielding a complexity of $O(k \cdot t_g)$. Integrating these components, the overall per-node complexity converges to
$O((s+q+k_c|\mathcal{P}|) \cdot t_g + t_m)$,
demonstrating that the complexity of IMDGA increases slowly with the scale of the dataset, being primarily constrained by the underlying Graph-LLMs and the chosen sampling hyperparameters.

\section{Experiments}

\definecolor{wwwgreen}{HTML}{253926}
\definecolor{wwwlise}{RGB}{244,234,197}
\definecolor{wwwpurple}{RGB}{159,137,158}
\definecolor{wwwpurple2}{RGB}{159,137,158}
\definecolor{wwwgreen2}{HTML}{83C896}
\definecolor{wwwgreen2}{HTML}{253926}
\definecolor{wwwgreen3}{HTML}{253926}
\definecolor{greenbox}{HTML}{C8EADE}
\definecolor{greenbox2}{HTML}{D1EBD6}
\definecolor{myred}{HTML}{8E0F31}
\definecolor{myblue}{HTML}{066190}
\definecolor{mygray}{HTML}{E1E1E1}
In this section, we conduct a comprehensive evaluation of the proposed IMDGA framework to demonstrate its effectiveness, interpretability, stealthiness, and robustness on TAGs. To rigorously assess these attributes, we structure our experiments around following research questions: \textbf{Q1:} To what extent does IMDGA method surpass traditional approaches in executing effective attacks on TAGs?  \textbf{Q2:} Can our method optimally reconcile the trade-off between attack stealthiness and effectiveness? \textbf{Q3:} Does IMDGA maintain robust attack performance when facing different Graph-LLM architectures? \textbf{Q4:} What distinct roles do the individual modules play in bolstering the overall performance of the IMDGA framework? \textbf{Q5:} How significantly do critical hyperparameters influence the robustness and adaptability of our attack strategy on TAGs?


\begin{table*}[t]
\caption{Comprehensive Comparison of Adversarial Attack Effectiveness Across TAG Datasets.}
\vspace{1mm}
\centering
\fontsize{9pt}{10pt}\selectfont
\renewcommand{\arraystretch}{1.1}
\resizebox{1\textwidth}{!}{
\begin{tabular}{llcccccccccc}
\arrayrulecolor{wwwgreen}\toprule[0.16em]
\multirow{2}{*}{\textbf{Dataset}} & \multirow{2}{*}{\textbf{Methods}} & \multicolumn{2}{c}{\textbf{SBERT}} & \multicolumn{2}{c}{\textbf{BERT}} & \multicolumn{2}{c}{\textbf{RoBERTa}} & \multicolumn{2}{c}{\textbf{DeBERTa}} & \multicolumn{2}{c}{\textbf{DistilBERT}}\\ 
\arrayrulecolor{wwwgreen2}\cmidrule[0.12em]{3-12} 
& & ACC \color{myblue}$\Downarrow$ & ASR  \color{myred} $\Uparrow$ & ACC\color{myblue}$\Downarrow$ & ASR \color{myred} $\Uparrow$ & ACC \color{myblue}$\Downarrow$ & ASR \color{myred} $\Uparrow$& ACC\color{myblue}$\Downarrow$ & ASR  \color{myred} $\Uparrow$& ACC \color{myblue}$\Downarrow$ & ASR\color{myred} $\Uparrow$\\ 
\arrayrulecolor{wwwgreen2}\midrule[0.08em] \addlinespace[-1.7pt]
\arrayrulecolor{wwwgreen3} \midrule[0.08em]
\multirow{7}{*}{\textbf{Cora}} 
& Clean 
& {82.14{\scriptsize \(\pm\)1.71}} 
& - 
& {80.61{\scriptsize \(\pm\)1.85}}
& - 
& {77.48{\scriptsize \(\pm\)0.91}} 
& - 
& {76.19{\scriptsize \(\pm\)1.16}} 
& - 
& {81.31{\scriptsize \(\pm\)0.06}} 
& -
\\
& HLBB 
& {76.97{\scriptsize \(\pm\)0.98}} 
& {23.33{\scriptsize \(\pm\)0.53}} 
& {76.42{\scriptsize \(\pm\)0.86}} 
& {32.41{\scriptsize \(\pm\)1.94}} 
& {73.70{\scriptsize \(\pm\)1.05}} 
& {24.54{\scriptsize \(\pm\)0.36}} 
& {74.80{\scriptsize \(\pm\)1.06}} 
& {9.72{\scriptsize \(\pm\)1.07}} 
& {78.22{\scriptsize \(\pm\)1.15}} 
& {23.15{\scriptsize \(\pm\)0.78}}
 \\
& TextHoaxer 
& {78.45{\scriptsize \(\pm\)1.35}}
& {16.68{\scriptsize \(\pm\)0.61}}
& {76.42{\scriptsize \(\pm\)1.92}}
& {32.41{\scriptsize \(\pm\)0.08}}
& {73.70{\scriptsize \(\pm\)1.17}}
& {24.54{\scriptsize \(\pm\)0.55}}
& {74.80{\scriptsize \(\pm\)1.04}}
& {9.72{\scriptsize \(\pm\)1.88}}
& {78.22{\scriptsize \(\pm\)0.33}}
& {23.15{\scriptsize \(\pm\)1.49}} \\
& SemAttack 
& {81.13{\scriptsize \(\pm\)0.29}}
& {4.44{\scriptsize \(\pm\)0.15}}
& {79.42{\scriptsize \(\pm\)0.91}}
& {6.05{\scriptsize \(\pm\)1.14}}
& {76.65{\scriptsize \(\pm\)0.58}}
& {5.09{\scriptsize \(\pm\)0.13}}
& {75.36{\scriptsize \(\pm\)0.36}}
& {6.48{\scriptsize \(\pm\)1.07}}
& {80.30{\scriptsize \(\pm\)1.96}}
& {6.48{\scriptsize \(\pm\)0.43}}\\
& FGSM
& 71.02 {\scriptsize \(\pm\)1.63}
& 43.70{\scriptsize \(\pm\)0.49}
& 70.37{\scriptsize \(\pm\)1.97}
& 57.21{\scriptsize \(\pm\)0.11}
& 64.61{\scriptsize \(\pm\)1.42}
& 63.43{\scriptsize \(\pm\)0.78}
& 69.17{\scriptsize \(\pm\)0.25}
& 45.37{\scriptsize \(\pm\)1.51}
& 70.47{\scriptsize \(\pm\)0.89}
& 61.11{\scriptsize \(\pm\)1.06} \\

& PA-F 
& 63.68 {\scriptsize \(\pm\)1.22}
& 70.74{\scriptsize \(\pm\)0.76}
& 71.94{\scriptsize \(\pm\)1.91}
& 46.05{\scriptsize \(\pm\)0.37}
& 64.28{\scriptsize \(\pm\)0.85}
& 61.11{\scriptsize \(\pm\)1.45}
& 67.60{\scriptsize \(\pm\)0.04}
& 48.61{\scriptsize \(\pm\)1.19}
& 75.50{\scriptsize \(\pm\)1.58}
& 37.96{\scriptsize \(\pm\)0.52} \\
& \cellcolor{mygray}IMDGA 
& {\textbf{61.51{\scriptsize \(\pm\)1.73}}}
& {\textbf{95.19{\scriptsize \(\pm\)0.02}}}
& {\textbf{64.33{\scriptsize \(\pm\)1.48}}}
& {\textbf{94.42{\scriptsize \(\pm\)0.87}}}
& {\textbf{61.98{\scriptsize \(\pm\)0.55}}}
& {\textbf{94.44{\scriptsize \(\pm\)1.94}}}
& {\textbf{61.24{\scriptsize \(\pm\)0.33}}}
& {\textbf{95.83{\scriptsize \(\pm\)1.21}}}
& {\textbf{65.25{\scriptsize \(\pm\)0.69}}}
& {\textbf{93.06{\scriptsize \(\pm\)1.08}}}
\\
\arrayrulecolor{wwwgreen2}\midrule[0.08em] \addlinespace[-1.7pt]
\arrayrulecolor{wwwgreen3} \midrule[0.08em]
\multirow{7}{*}{\textbf{Citeseer}} 
& Clean 
& {71.58{\scriptsize \(\pm\)1.16}}
& -
& {72.54{\scriptsize \(\pm\)0.50}}
& {-}
& {71.48{\scriptsize \(\pm\)0.82}}
& {-}
& {70.26{\scriptsize \(\pm\)0.91}}
& {-}
& {72.85{\scriptsize \(\pm\)0.37}}
& {-}\\
& HLBB 
& {64.14{\scriptsize \(\pm\)1.05}}
& {43.08{\scriptsize \(\pm\)0.67}}
& {67.28{\scriptsize \(\pm\)1.93}}
& {40.16{\scriptsize \(\pm\)0.18}}
& {69.16{\scriptsize \(\pm\)1.34}}
& {18.50{\scriptsize \(\pm\)0.42}}
& {66.42{\scriptsize \(\pm\)0.27}}
& {26.38{\scriptsize \(\pm\)1.58}}
& {69.48{\scriptsize \(\pm\)1.77}}
& {29.13{\scriptsize \(\pm\)0.96}}\\
& TextHoaxer
& {66.65{\scriptsize \(\pm\)1.41}}
& {27.36{\scriptsize \(\pm\)0.98}}
& {69.75{\scriptsize \(\pm\)0.25}}
& {23.23{\scriptsize \(\pm\)1.80}}
& {69.83{\scriptsize \(\pm\)0.07}}
& {14.57{\scriptsize \(\pm\)1.63}}
& {68.73{\scriptsize \(\pm\)0.56}}
& {11.02{\scriptsize \(\pm\)1.97}}
& {70.46{\scriptsize \(\pm\)0.39}}
& {20.47{\scriptsize \(\pm\)1.11}}\\
& SemAttack
& {70.26{\scriptsize \(\pm\)1.25}}
& {7.23{\scriptsize \(\pm\)0.78}}
& {71.32{\scriptsize \(\pm\)1.91}}
& {8.66{\scriptsize \(\pm\)0.33}}
& {71.17{\scriptsize \(\pm\)0.62}}
& {3.15{\scriptsize \(\pm\)1.06}}
& {68.07{\scriptsize \(\pm\)1.54}}
& {15.35{\scriptsize \(\pm\)0.09}}
& {71.95{\scriptsize \(\pm\)1.77}}
& {9.06{\scriptsize \(\pm\)0.48}}\\
& FGSM
& 59.83 {\scriptsize \(\pm\)1.04}
& 58.81{\scriptsize \(\pm\)0.61}
& 64.26{\scriptsize \(\pm\)1.98}
& 51.97{\scriptsize \(\pm\)0.15}
& 61.91{\scriptsize \(\pm\)1.42}
& 61.02{\scriptsize \(\pm\)0.73}
& 63.52{\scriptsize \(\pm\)0.36}
& 43.31{\scriptsize \(\pm\)1.55}
& 64.73{\scriptsize \(\pm\)0.89}
& 55.51{\scriptsize \(\pm\)1.22}\\
& PA-F 
& 68.77{\scriptsize \(\pm\)1.33}
& 14.15{\scriptsize \(\pm\)0.69}
& 69.95{\scriptsize \(\pm\)1.92}
& 19.29{\scriptsize \(\pm\)0.18}
& 62.18{\scriptsize \(\pm\)1.04}
& 58.27{\scriptsize \(\pm\)0.77}
& 61.75{\scriptsize \(\pm\)0.25}
& 55.51{\scriptsize \(\pm\)1.55}
& 69.16{\scriptsize \(\pm\)0.48}
& 27.95{\scriptsize \(\pm\)1.87} 
\\
& \cellcolor{mygray}{IMDGA} 
& {\textbf{53.82{\scriptsize \(\pm\)1.04}}}
& {\textbf{92.77{\scriptsize \(\pm\)0.61}}}
& {\textbf{58.65{\scriptsize \(\pm\)1.98}}}
& {\textbf{94.49{\scriptsize \(\pm\)0.15}}}
& {\textbf{57.94{\scriptsize \(\pm\)1.42}}}
& {\textbf{93.31{\scriptsize \(\pm\)0.73}}}
& {\textbf{56.89{\scriptsize \(\pm\)0.36}}}
& {\textbf{93.31{\scriptsize \(\pm\)1.55}}}
& {\textbf{59.40{\scriptsize \(\pm\)0.89}}}
& {\textbf{94.10{\scriptsize \(\pm\)1.22}}}
\\
\arrayrulecolor{wwwgreen2}\midrule[0.08em] \addlinespace[-1.7pt]
\arrayrulecolor{wwwgreen3} \midrule[0.08em]
\multirow{7}{*}{\textbf{Pubmed}} 
& Clean 
& {82.31{\scriptsize \(\pm\)1.45}}
& {-}
&{82.31{\scriptsize \(\pm\)0.78}}
& {-}
& {83.71{\scriptsize \(\pm\)0.29}}
& {-}
& {82.62{\scriptsize \(\pm\)1.94}}
& {-}
& {82.04{\scriptsize \(\pm\)0.05}}
& {-}\\
& HLBB 
& {81.12{\scriptsize \(\pm\)1.73}}
& {27.67{\scriptsize \(\pm\)0.02}}
& {81.25{\scriptsize \(\pm\)1.48}}
& {30.00{\scriptsize \(\pm\)0.87}}
& {82.95{\scriptsize \(\pm\)0.55}}
& {25.80{\scriptsize \(\pm\)1.94}}
& {82.03{\scriptsize \(\pm\)0.33}}
& {18.60{\scriptsize \(\pm\)1.21}}
& {81.29{\scriptsize \(\pm\)0.69}}
& {23.80{\scriptsize \(\pm\)1.08}}\\
& TextHoaxer 
& {81.15{\scriptsize \(\pm\)1.19}}
& {28.67{\scriptsize \(\pm\)0.58}}
& {81.27{\scriptsize \(\pm\)1.92}}
& {28.20{\scriptsize \(\pm\)0.03}}
& {82.88{\scriptsize \(\pm\)1.35}}
& {25.00{\scriptsize \(\pm\)0.81}}
& {82.03{\scriptsize \(\pm\)0.46}}
& {16.80{\scriptsize \(\pm\)1.64}}
& {81.33{\scriptsize \(\pm\)0.99}}
& {22.00{\scriptsize \(\pm\)1.87}}\\
& SemAttack
& {81.93{\scriptsize \(\pm\)1.04}}
& {16.00{\scriptsize \(\pm\)0.61}}
& {82.10{\scriptsize \(\pm\)1.98}}
& {5.40{\scriptsize \(\pm\)0.15}}
& {83.54{\scriptsize \(\pm\)1.42}}
& {5.40{\scriptsize \(\pm\)0.73}}
& {82.43{\scriptsize \(\pm\)0.36}}
& {17.20{\scriptsize \(\pm\)1.55}}
& {81.97{\scriptsize \(\pm\)0.89}}
& {4.00{\scriptsize \(\pm\)0.22}}\\

& FGSM
& 80.27{\scriptsize \(\pm\)1.19}
& 64.33{\scriptsize \(\pm\)0.58}
& 78.65 {\scriptsize \(\pm\)1.92}
& 69.00{\scriptsize \(\pm\)0.03}
& 80.67{\scriptsize \(\pm\)1.35}
& 72.40{\scriptsize \(\pm\)0.81}
& 81.67{\scriptsize \(\pm\)0.46}
& 35.20{\scriptsize \(\pm\)1.64}
& 78.59{\scriptsize \(\pm\)0.99}
& 57.40{\scriptsize \(\pm\)1.87} \\

& PA-F 
& 80.20{\scriptsize \(\pm\)1.04}
& 13.29{\scriptsize \(\pm\)0.61}
& 80.76{\scriptsize \(\pm\)1.98}
& 29.40{\scriptsize \(\pm\)0.15}
& 80.65{\scriptsize \(\pm\)1.42}
& 51.00{\scriptsize \(\pm\)0.73}
& 81.70{\scriptsize \(\pm\)0.36}
& 32.60{\scriptsize \(\pm\)1.55}
& 80.67{\scriptsize \(\pm\)0.89}
& 17.80{\scriptsize \(\pm\)1.22} 
\\
& \cellcolor{mygray}{IMDGA} 
& {\textbf{77.62{\scriptsize \(\pm\)1.19}}}
& {\textbf{85.88{\scriptsize \(\pm\)0.58}}}
& {\textbf{77.84{\scriptsize \(\pm\)1.92}}}
& {\textbf{86.40{\scriptsize \(\pm\)0.03}}}
& {\textbf{80.17{\scriptsize \(\pm\)1.35}}}
& {\textbf{84.40{\scriptsize \(\pm\)0.81}}}
& {\textbf{79.35{\scriptsize \(\pm\)0.46}}}
& {\textbf{77.40{\scriptsize \(\pm\)1.64}}}
& {\textbf{77.88{\scriptsize \(\pm\)0.99}}}
& {\textbf{86.00{\scriptsize \(\pm\)1.87}}}
\\

\arrayrulecolor{wwwgreen}\bottomrule[0.16em]
\end{tabular}
}
\label{table1}
\end{table*}

\subsection{Experimental Setup}
\textbf{Datasets.} 
We evaluate our proposed method on several widely used benchmark datasets. Specifically, we adopt Cora, Citeseer, and PubMed~\cite{Yang16cora}, which are among the most frequently used citation datasets. Beyond these commonly used datasets, we further incorporate the large-scale ogbn-arxiv~\cite{hu2020ogb} dataset to assess the scalability and practicality of our approach in more challenging, real-world scenarios. Detailed dataset statistics are provided in Appendix~\ref{app: datasets}.\\
\textbf{Compared Baselines.} 
To ensure a fair and comprehensive comparison of adversarial performance on TAGs, we select baselines from both the text adversarial attack domain and the graph adversarial attack domain (see Appendix \ref{app:baseline}). From the former category, we select HLBB~\cite{mu2021hard}, TextHoaxer~\cite{ye2022texthoaxer}, and SemAttack~\cite{wang2022semattack}, adapting them to TAGs to evaluate their transferability from text-only tasks to TAGs settings. From the latter, we adopt FGSM~\cite{goodfellow2014explaining} and PA-F~\cite{ma2020towards}, two representative approaches in graph adversarial attack, which enable a direct evaluation against graph-specific perturbations.\\
\textbf{Evaluation.}
Since our attack primarily targets the node classification task, we evaluate effectiveness from both a global perspective and a local perspective. From the global view, we measure the overall classification ACC on the test set, which reflects how the attack influences the general predictive capability of the victim model. From the local view, we adopt the ASR on originally correctly classified nodes, providing a more fine-grained measure of how effectively the attack disrupts predictions at the target node. For more detailed experimental settings, refer to the Appendix \ref{app: exp settings}. 

\subsection{Performance Comparison (Q1)}
Our experimental results consistently demonstrate the superior effectiveness and robustness of the proposed method across both local and global perspectives, substantially outperforming existing baselines and exposing the vulnerability of Graph-LLMs to adversarial manipulations. Through comprehensive data analysis, we derive the following key observations: (1) \textbf{Comprehensive SOTA performance:} At the local level, our strategy consistently surpasses all competing methods, producing the strongest disruptive effect on target nodes across every dataset and Graph-LLM variant. On smaller benchmarks such as Cora and Citeseer, the ASR exceeds 90\%, showing that nearly all targeted nodes can be reliably compromised. Even on larger-scale datasets like ogbn-arxiv, where adversarial robustness is typically higher, our method still demonstrates clear superiority over existing baselines. From the global perspective, although the degree of degradation is influenced by factors such as node selection strategy and perturbation scale, under a unified random selection protocol our approach still achieves significantly stronger global disruption than any baseline. 
(2) \textbf{Overcoming prior limitations:} Traditional NLP adversarial attacks, despite their success in text classification, exhibit severely diminished transferability in the context of TAGs. Even in its best case, HLBB achieves only 40.16\% ASR on Citeseer (with BERT encoding), while in other scenarios certain methods yield ASR values dropping below 10\%. Meanwhile, graph-specific adversarial baselines such as FGSM and PA-F perform better under the same similarity constraints, but remain limited to embedding-level manipulations without semantic interpretability. By contrast, our method not only achieves stronger quantitative performance but also introduces meaningful and explainable perturbations, thereby addressing the twofold shortcomings of insufficient transferability in text-based attacks and lack of interpretability in graph-based ones. 

\begin{figure*}[ht]
    \centering
    \begin{minipage}{0.48\textwidth}
        \centering
        \begin{minipage}{0.32\linewidth}
            \centering
            \includegraphics[width=\linewidth]{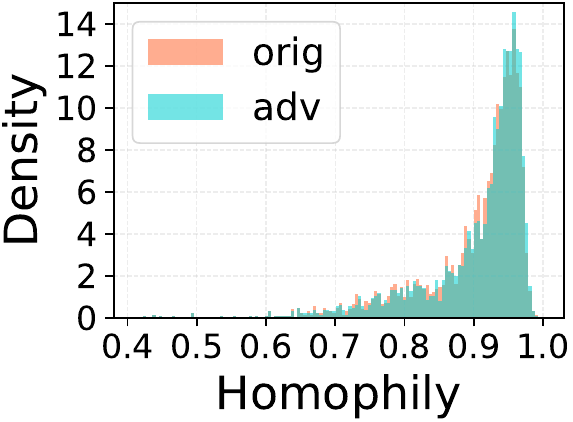}\\[-2pt]
            \small (a) Cora
        \end{minipage}\hfill
        \begin{minipage}{0.32\linewidth}
            \centering
            \includegraphics[width=\linewidth]{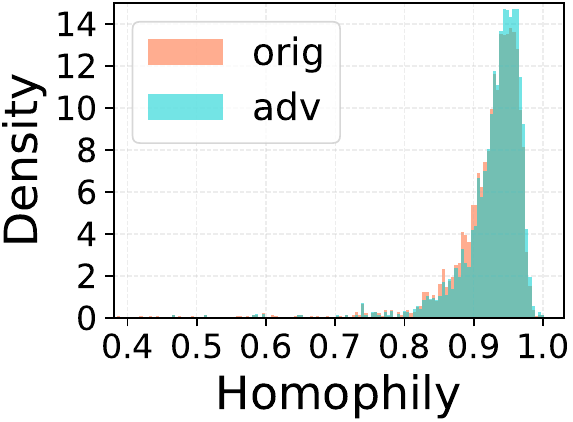}\\[-2pt]
            \small (b) Citeseer
        \end{minipage}\hfill
        \begin{minipage}{0.32\linewidth}
            \centering
            \includegraphics[width=\linewidth]{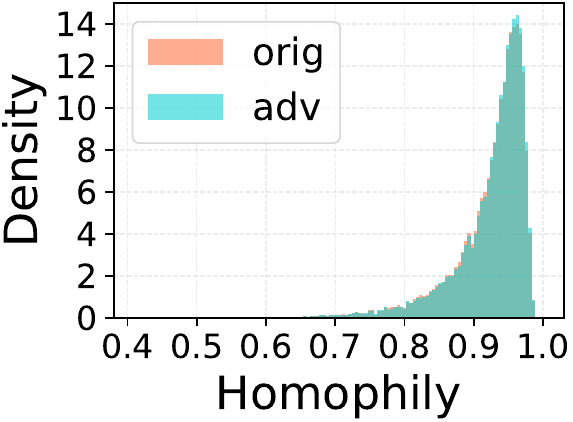}\\[-2pt]
            \small (c) PubMed
        \end{minipage}
        \caption{Feature-cosine homophily (before and after attack).}
        \label{fig:homophily_triplet}
    \end{minipage}\hfill
    \begin{minipage}{0.48\textwidth}
        \centering
        \begin{minipage}{0.32\linewidth}
            \centering
            \includegraphics[width=\linewidth]{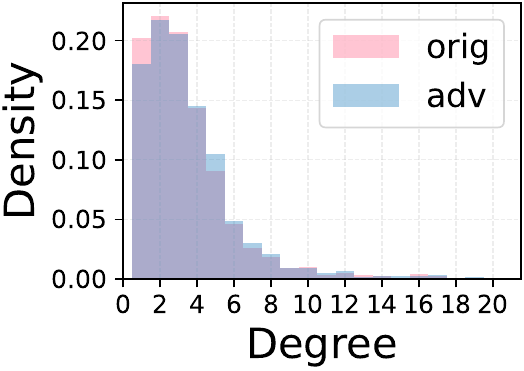}\\[-2pt]
            \small (a) Cora
        \end{minipage}\hfill
        \begin{minipage}{0.32\linewidth}
            \centering
            \includegraphics[width=\linewidth]{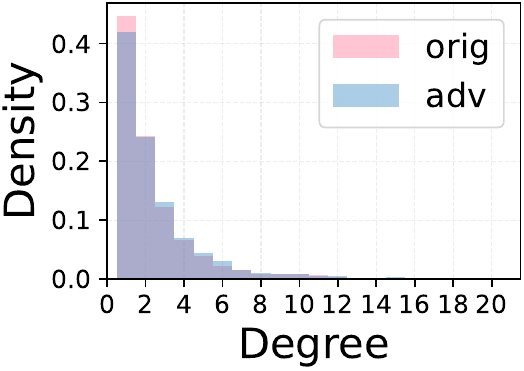}\\[-2pt]
            \small (b) Citeseer
        \end{minipage}\hfill
        \begin{minipage}{0.32\linewidth}
            \centering
            \includegraphics[width=\linewidth]{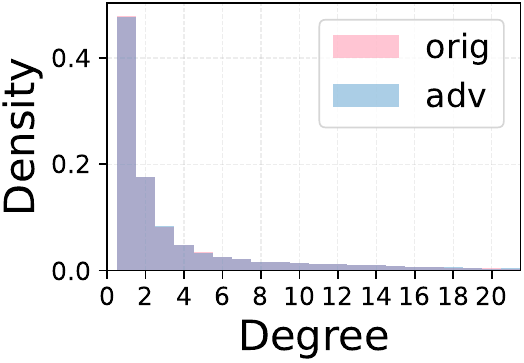}\\[-2pt]
            \small (c) PubMed
        \end{minipage}
        \caption{Degree distributions (before and after attack).}
        \label{fig:degree_triplet}
    \end{minipage}
\end{figure*}

\subsection{Stealthiness Evaluation (Q2)}
\label{subsec:q2_stealth}


In this section, we evaluate stealthiness from two perspectives aligned with IMDGA's modules. For the textual perspective, we select BERT cosine similarity (Sim), GPT-2 perplexity (PPL)~\cite{radford2019language}, and human ratings (see Appendix~\ref{app:human}) as our main metrics.
From the structural perspective, we report both degree distributions and the feature-cosine homophily.
Specifically, the feature-cosine homophily of each node $u$ is formally defined as:
\begin{equation}
h_u = \cos(r_u, x_u), \qquad 
r_u = \sum_{j \in \mathcal N(u)} \frac{1}{\sqrt{d_u d_j}}\, x_j,
\label{eq:feature_cosine_homophily}
\end{equation}
where $\mathcal N(u)$ denotes the immediate neighbor set of $u$, $d_u$ indicates its structural degree, and $x_u$ characterizes its node feature vector. In particular, $r_u$ represents the normalized aggregation of neighbor features. 
Structural stealthiness is measured by $\Delta h_u$ and $\Delta d_u$, where smaller deviations imply higher stealthiness.

Under the same perturbation ratio, IMDGA consistently achieves higher Sim and lower PPL than text-only baselines, as clearly shown in Table~\ref{tab:stealth_text}, which is consistent with human ratings and further indicates better semantic preservation and fluency.

The \emph{Semantic Substitution Module} explains these improvements. 
Eq.~(11) employs an MLM to propose context-aware candidates, while Eqs.~(12)–(14) evaluate replacements via neighborhood confidence gaps with an adaptive label-flip factor.
If a fraction $\rho$ of tokens is replaced with candidates under an MLM similarity threshold $\gamma$, then with $\ell_2$-normalized embeddings we obtain:
\begin{equation}
\|h'-h\|\le \rho\sqrt{2(1-\gamma)} \;\Rightarrow\; \cos(h',h)\ge 1-\rho^2(1-\gamma).
\label{eq:simlower}
\end{equation}
This inequality establishes a lower bound on similarity with small $\rho$ and large $\gamma$.
Moreover, replacements with sufficiently high relative likelihood incur only a limited increase in negative log-likelihood (NLL), hence a mild rise in PPL. 
Together with the adaptive scoring in Eq.~(14), module favors substitutions that are both impactful and linguistically plausible, yielding effective yet stealthy perturbations.



The homophily distributions after the \emph{Edge Pruning Module} almost coincide with the originals, with minor tail deviations, as shown in Figure~\ref{fig:homophily_triplet}. 
This arises from pruning high-attribution edges with a small budget (top-$k$ in Eq.~(18)) and degree-normalized aggregation in Eq.~\eqref{eq:feature_cosine_homophily}. 
Deleting an edge $(u,v)$ updates $r_u'=r_u-\tfrac{x_v}{\sqrt{d_ud_v}}$, and for unit-normalized features a first-order bound gives:
\begin{equation}
\big|\cos(r_u',x_u)-\cos(r_u,x_u)\big|\;\lesssim\;\frac{1}{\sqrt{d_ud_v}}\cdot\frac{\|x_v\|}{\|r_u\|},
\label{eq:homobound}
\end{equation}
so the change in $h_u$ is constrained by degree normalization and the limited number of pruned edges.


The degree distributions, as clearly shown in Figure~\ref{fig:degree_triplet}, preserve the head and body of the original curves with only negligible tail differences, thereby indicating that the global topology is statistically indistinguishable from the original. 
Let $|\Delta E|$ be the number of pruned edges. 
Since each pruned edge decreases the degrees of its two endpoints by one, we have:
\begin{equation}
\frac{1}{|V|}\sum_{u}|d'_u-d_u| \;=\; \frac{2|\Delta E|}{|V|},
\label{eq:degreebound}
\end{equation}
and node-wise caps together with the locality of the nexus $G_n(v)$ (Eq.~(17)) keep $|\Delta E|/|V|$ small, explaining the near overlap.


\begin{table}[!t]
    \centering
    \setlength{\abovecaptionskip}{0.2cm}
    \setlength{\belowcaptionskip}{-0.2cm}
    \caption{Textual stealthiness.}
    \label{tab:stealth_text}
    \arrayrulecolor{wwwgreen}
    \begin{tabularx}{\linewidth}{@{\hspace{4pt}} l
      Y@{\hspace{2pt}}Y@{\hspace{2pt}}Y
      Y@{\hspace{2pt}}Y@{\hspace{2pt}}Y @{\hspace{4pt}}}
        \toprule[0.16em]
        \multirow{2}{*}{Method}
        & \multicolumn{3}{@{\hspace{2pt}}c@{\hspace{2pt}}}{Cora}
        & \multicolumn{3}{@{\hspace{2pt}}c@{\hspace{2pt}}}{Citeseer} \\
        \cmidrule(l{4pt}r{4pt}){2-4}\cmidrule(l{4pt}r{4pt}){5-7}
         & PPL & Sim & Human
         & PPL & Sim & Human \\
        \midrule[0.16em]
        HLBB        & 638.35 & 0.917 & 3.254 & 574.46 & 0.934 & 3.391 \\
        TextHoaxer  & 618.55 & 0.914 & 3.290 & 580.38 & 0.955 & 3.326 \\
        SemAttack   & 498.46 & 0.954 & 3.486 & 209.99  & 0.958 & 3.493 \\
        \cellcolor{mygray}IMDGA & \textbf{196.74} & \textbf{0.955} & \textbf{3.565} & \textbf{186.18} & \textbf{0.960} & \textbf{3.529} \\
        \bottomrule[0.16em]
    \end{tabularx}
    \arrayrulecolor{black}
\end{table}

\begin{table}[H]
    \setlength{\abovecaptionskip}{0.2cm}
    \setlength{\belowcaptionskip}{-0.2cm}
    \caption{Generalizable Robustness and Effectiveness of IMDGA Across Multiple Backbones (ASR).}
    \label{tab::robustness}
    \centering
    \begin{tabular}{ccccc}
    \arrayrulecolor{wwwgreen}\toprule[0.16em]
        Encoder & {Backbone} & Cora & Citeseer & PubMed  \\ 
    \arrayrulecolor{wwwgreen} \midrule[0.16em]
        \multirow{7}{*}{SBERT} 
        & GCN & 92.19 ± 0.84 & 94.87 ± 0.42 & 87.22 ± 1.02  \\
        & GAT & 83.58 ± 0.67 & 93.48 ± 1.35 & \textbf{90.72 ± 0.64}  \\ 
        & SAGE & \textbf{95.21 ± 1.89} & 92.74 ± 0.58 & 88.79 ± 1.05  \\
        & RGCN & 89.04 ± 0.35 & 91.19 ± 1.81 & 84.66 ± 0.92  \\ 
        & Guard & 93.92 ± 1.76 & \textbf{98.12 ± 0.36} & 90.12 ± 1.58  \\
        & GLEM & 88.73 ± 0.54 & 94.55 ± 0.43 & 90.11 ± 0.47  \\
        & GIANT & 87.19 ± 0.98 & 93.67 ± 1.24 & 84.39 ± 1.26 \\
        \arrayrulecolor{wwwgreen} \midrule[0.08em]
        \arrayrulecolor{wwwgreen} \midrule[0.08em]
                \multirow{7}{*}{BERT} 
        & GCN & \textbf{92.37 ± 1.42} & 95.63 ± 0.72 & 85.42 ± 1.37  \\
        & GAT   & 90.12 ± 0.74 & 93.26 ± 1.18 & \textbf{91.73 ± 0.72}  \\
        & SAGE  & 89.77 ± 1.63 & 92.41 ± 0.66 & 89.04 ± 0.92  \\
        & RGCN  & 87.03 ± 0.42 & 91.86 ± 1.64 & 85.13 ± 1.08  \\
        & Guard & 86.47 ± 1.59 & \textbf{97.89 ± 0.41} & 91.28 ± 1.33  \\
        & GLEM  & 89.21 ± 0.61 & 94.32 ± 0.49 & 90.54 ± 0.53  \\
        & GIANT & 87.68 ± 1.05 & 93.24 ± 1.36 & 84.92 ± 1.12  \\
    \arrayrulecolor{wwwgreen} \midrule[0.16em]
    \end{tabular}
\end{table}



\subsection{Attack Robustness Assessment (Q3)}
Having demonstrated the effectiveness and stealthiness of IMDGA in previous sections, we further investigate its generalization and robustness through extensive experiments. To verify that our attack is not limited to a specific GNN architecture, we evaluate IMDGA on widely-used models such as GCN~\cite{kipf2016gcn}, GAT~\cite{velivckovic2017gat}, and GraphSAGE~\cite{hamilton2017graphsage}, as well as on more robust architectures including RGCN~\cite{schlichtkrull2018modeling} and GNNGuard~\cite{zhang2020gnnguard}. For Graph-LLMs, although Q1 compared multiple encoder models, the attack performance on more advanced Graph-LLMs remains unexplored. Therefore, we conduct comprehensive and rigorous experiments using GLEM~\cite{zhao2023learning} and GIANT~\cite{chien2022node} as representative Graph-LLMs.

\begin{figure*}[htbp]
\centering
\begin{subfigure}[b]{0.24\textwidth}
    \includegraphics[width=\textwidth]{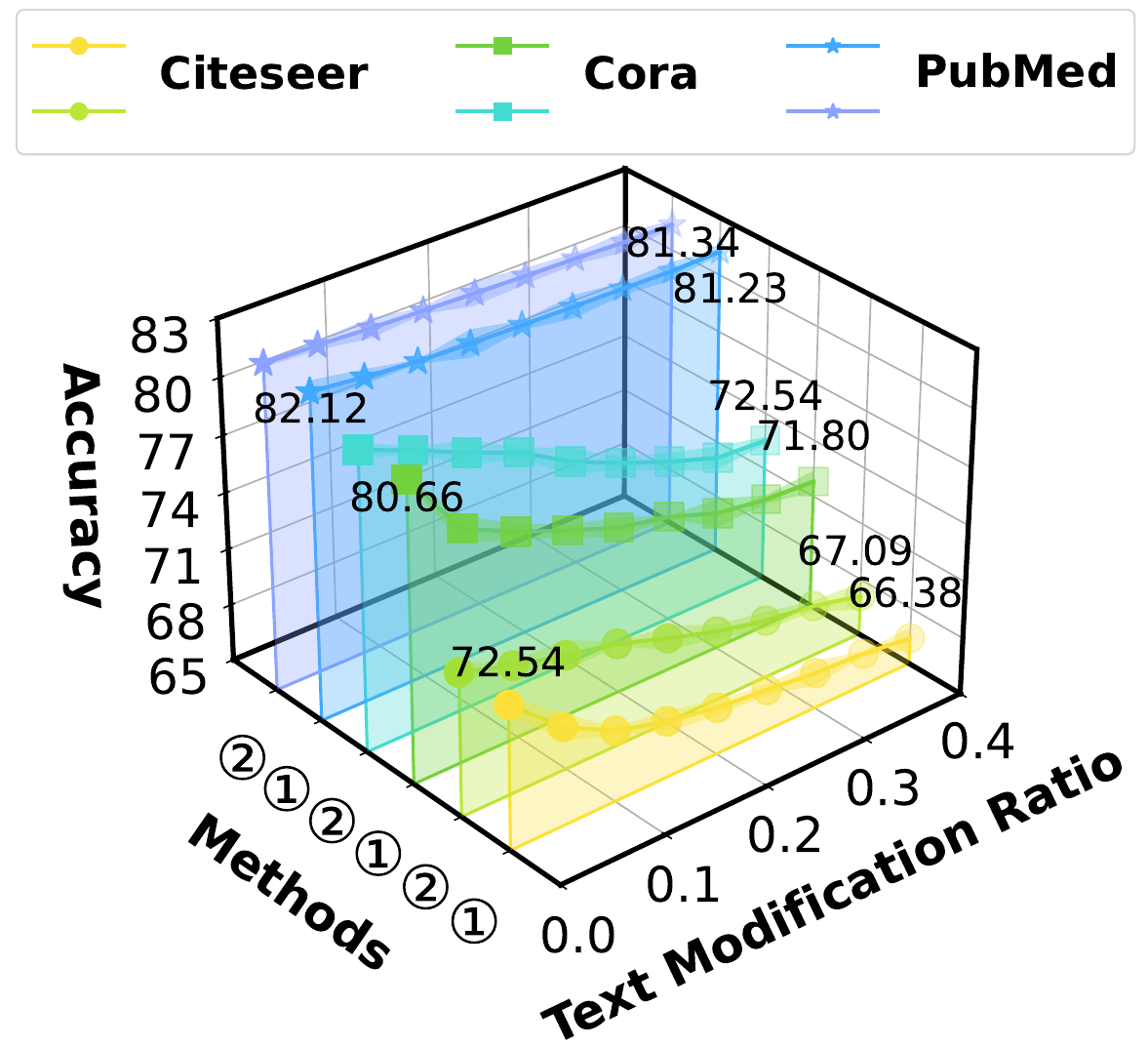}
\end{subfigure}
\hfill
\begin{subfigure}[b]{0.24\textwidth}
    \includegraphics[width=\textwidth]{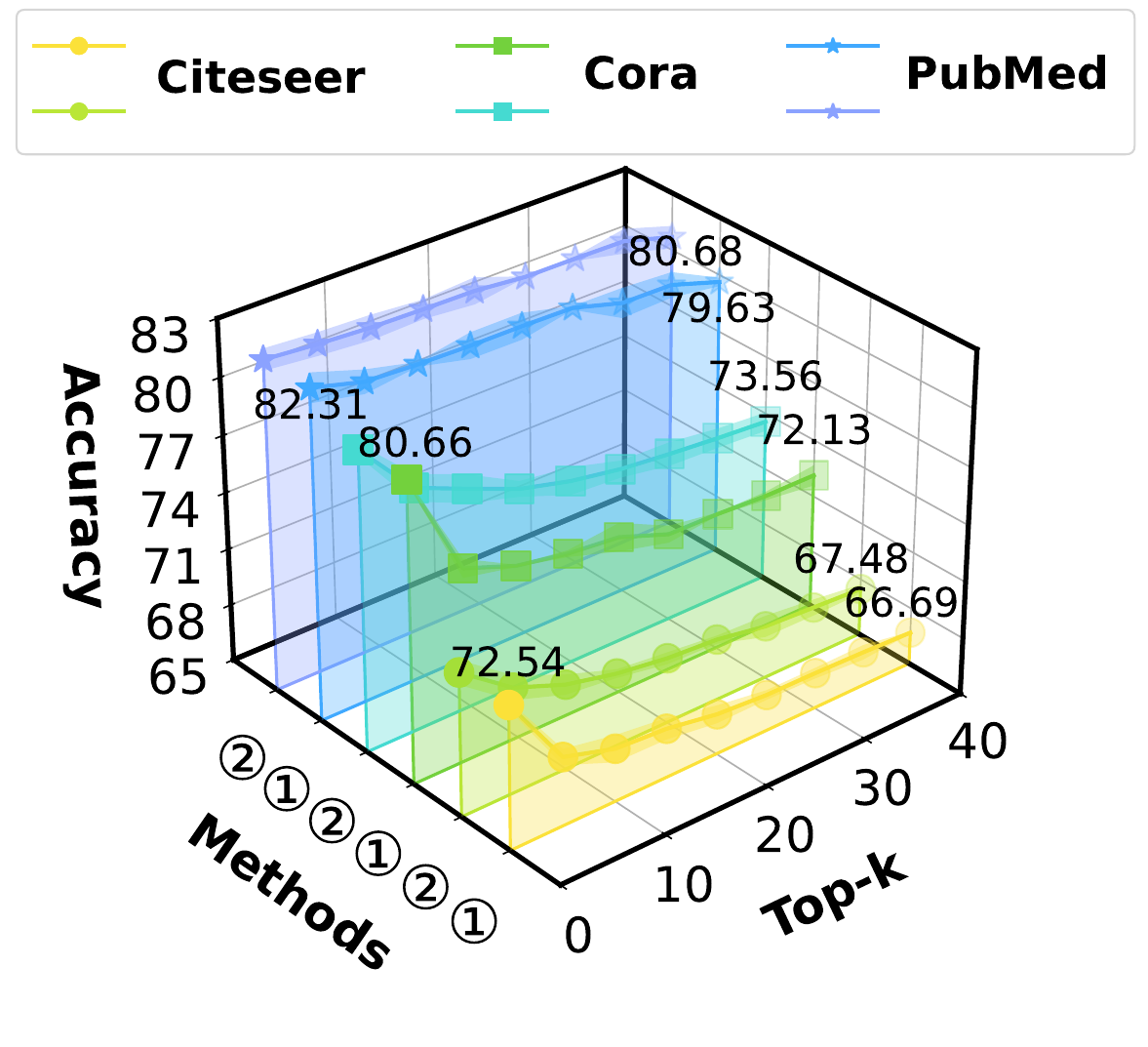}
\end{subfigure}
\hfill
\begin{subfigure}[b]{0.24\textwidth}
    \includegraphics[width=\textwidth]{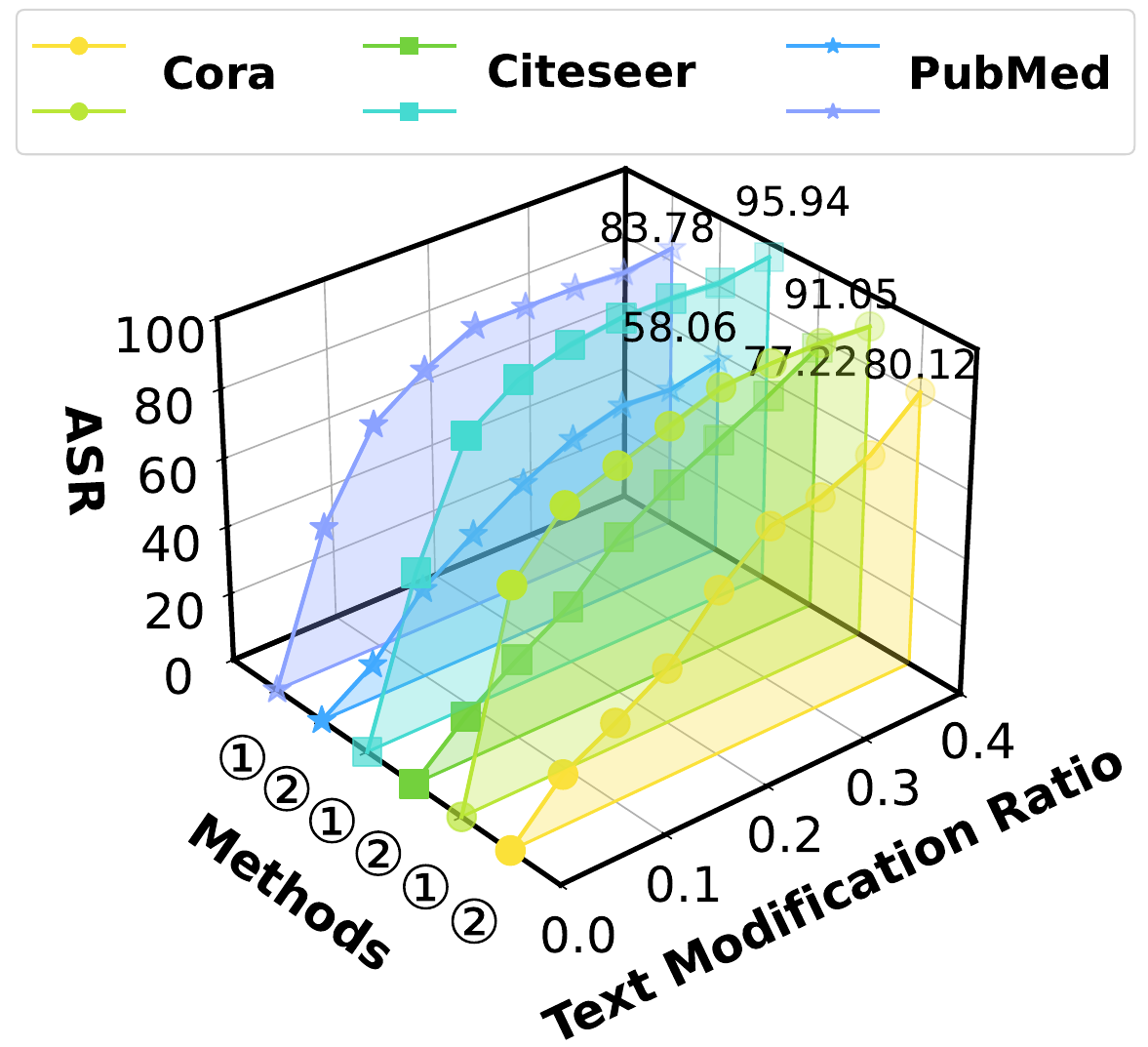}
\end{subfigure}
\hfill
\begin{subfigure}[b]{0.24\textwidth}
    \includegraphics[width=\textwidth]{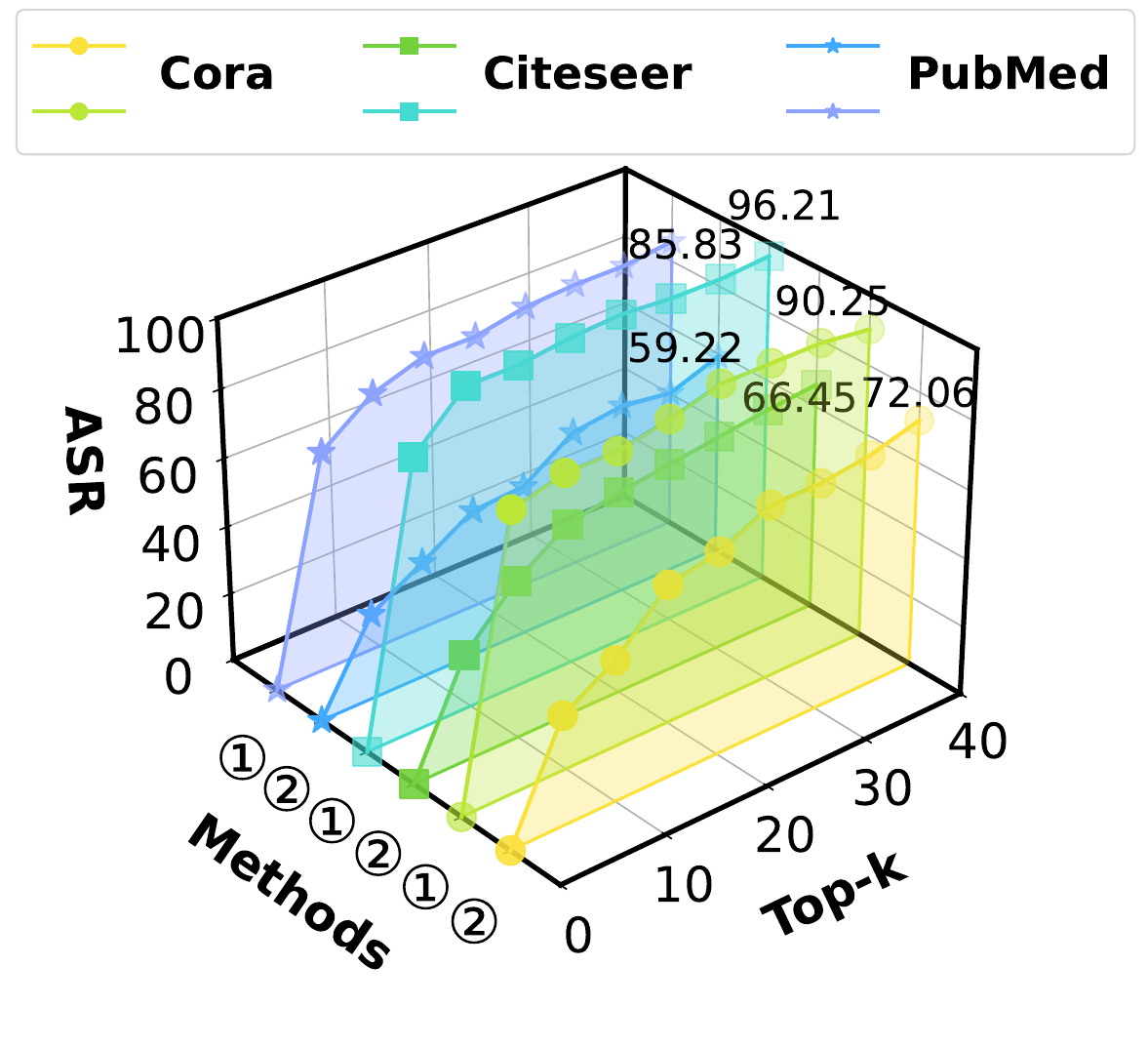}
\end{subfigure}
\caption{Sensitivity of IMDGA (\ding{172} represents the full IMDGA and \ding{173} denotes text-only perturbations).}
\label{fig:hyper}
\end{figure*}
The results presented in Table~\ref{tab::robustness} indicate that all tested models exhibit relatively weak resistance to IMDGA, with only minor differences in ASR across architectures. Notably, even the robust Guard model achieves an ASR of 98.12\% on Citeseer, suggesting that while these backbones leverage node similarity to enhance robustness, they remain vulnerable to attacks exploiting textual information. Similarly, GLEM and GIANT, which employ fine-tuned encoders to incorporate richer node representations, do not demonstrate improved resistance against our proposed adversarial attack. These findings collectively highlight that IMDGA consistently compromises diverse GNNs and Graph-LLMs, underscoring the persistent vulnerabilities present across modern graph-based models.

\subsection{Ablation Study (Q4)}
To evaluate the contribution of each component in our adversarial framework, we conduct ablation experiments and report the results in Table~\ref{tab::abltion}. For the Topological SHAP Module, we replace the identified pivotal words with randomly selected words for comparison. For the Semantic Perturbation Module, while it is infeasible to completely remove the effect of the MLM, we substitute the graph-aware scoring function originally designed to integrate both structural and textual information with a simpler function that only considers the prediction probability drop of a single node. For the Edge Pruning Module, we directly eliminate it, leaving only text-based perturbations in the algorithm. It is worth noting that, to preserve stealthiness, we did not apply the Edge Pruning strategy when ablating the first two modules.\\
From the analysis of the results, several conclusions can be drawn:
(1) Text modification modules are critical to attack success. When perturbations are restricted to naive random word substitutions or applied without incorporating graph structural information, the ASR sharply drops from around 92\% to below 40\%. Such simplified attacks are insufficient to substantially compromise the robustness of Graph-LLMs. By contrast, precisely identifying pivotal words in message passing and applying carefully designed graph-aware substitutions leads to a stronger adversarial influence on TAGs.\\
(2) Word-level perturbations alone face inherent limitations. Even when pivotal words are correctly identified and semantically reasonable substitutions are introduced, the ASR remains capped at around 60\%. Without the complementary structural guidance provided by edge-level manipulations, attacks struggle to overcome the inherent robustness of representation learning in TAGs.

\subsection{Hyperparameter Analysis (Q5)}
To address Q5, we conducted extensive experiments focusing on two pivotal hyperparameters: the text modification ratio $\beta$ and the top-$k$ selection parameter. These two factors largely govern the strength of adversarial perturbations and thus serve as the most representative indicators for sensitivity analysis. In contrast, other hyperparameters play a more secondary role and are discussed in detail, along with their corresponding search spaces, in Appendix \ref{app:hypersetting}. For experimental efficiency and comparability, we randomly selected 100 nodes as target instances in each trial, ensuring stable and reliable evaluation across different settings.\\
The results, summarized in Figure~\ref{fig:hyper}, reveal a consistent and interpretable trend: as $\beta$ and top-$k$ increase, both ACC and ASR undergo a rapid initial change, followed by a clear attenuation in marginal gains. This observation suggests that moderate relaxation of these constraints brings substantial improvements in attack effectiveness at first, as models quickly become more vulnerable under stronger perturbations. However, further increases lead to diminishing returns, with only marginal benefits despite greater perturbation. More critically, loosening these hyperparameters significantly raises the risk of exposure, as excessive modifications are more detectable, while also incurring higher computational and temporal overhead. Consequently, there exists an inherent trade-off among effectiveness, stealth, and efficiency when choosing hyperparameters. Practically, this trade-off implies that optimal settings for $\beta$ and top-$k$ should be chosen according to the deployment scenario—prioritizing higher stealth for stealth-sensitive applications and higher strength where maximal disruption is required.

\begin{table} 
\setlength{\abovecaptionskip}{0.2cm}
\setlength{\belowcaptionskip}{-0.2cm}
\caption{Ablation study (ASR).
}
\label{tab::abltion}
\begin{tabular}{cccc}
\arrayrulecolor{wwwgreen}\toprule[0.16em]
{Model} & Cora  & Citeseer   & PubMed\\
\arrayrulecolor{wwwgreen} \midrule[0.16em]
IMDGA (SBERT)                                      
& 91.78±0.72 & 94.86±0.79 & 85.47±0.41\\
w/o Topology SHAP                       
& 29.54±0.42 & 32.89±0.29 & 23.76±0.58\\
w/o Semantic Perturb                     
& 36.97±0.53 & 41.21±0.19 & 32.84±0.44\\
w/o Edge Pruning                        
& 66.12±0.35 & 62.47±0.33 & 59.21±0.55\\  
\arrayrulecolor{wwwgreen} \midrule[0.08em]
\arrayrulecolor{wwwgreen} \midrule[0.08em]
IMDGA (BERT)                                      
& 92.06±0.61 & 95.11±0.87 & 85.16±0.35\\
w/o Topology SHAP                       
& 29.13±0.38 & 33.18±0.21 & 23.41±0.63\\
w/o Semantic Perturb                     
& 37.32±0.46 & 40.93±0.16 & 33.21±0.38\\
w/o Edge Pruning                        
& 65.87±0.30 & 62.23±0.26 & 58.83±0.62\\
\arrayrulecolor{wwwgreen} \midrule[0.16em]
\end{tabular}
\vspace{-0.3cm}
\end{table}

\section{Conclusion}
In this work, we introduced IMDGA to investigate the heightened vulnerabilities of TAGs induced by the integration of textual features into Graph-LLMs. By jointly leveraging Topological SHAP, Semantic Perturbation, and Edge Pruning, IMDGA orchestrates multi-layered adversarial manipulations that expose weaknesses in both textual and structural dimensions. Through extensive evaluations, our method consistently outperforms conventional NLP and graph adversarial attack baselines in terms of interpretability, effectiveness, and stealthiness, achieving high ASR across diverse datasets. These findings not only uncover fundamental fragilities in TAG representation learning but also underscore the urgent need for systematic defenses to guide the development of more resilient Graph-LLMs against increasingly sophisticated adversarial threats.

\clearpage
\bibliographystyle{ACM-Reference-Format}
\bibliography{ref.bib}


\begin{thebibliography}{72}


\ifx \showCODEN    \undefined \def \showCODEN     #1{\unskip}     \fi
\ifx \showISBNx    \undefined \def \showISBNx     #1{\unskip}     \fi
\ifx \showISBNxiii \undefined \def \showISBNxiii  #1{\unskip}     \fi
\ifx \showISSN     \undefined \def \showISSN      #1{\unskip}     \fi
\ifx \showLCCN     \undefined \def \showLCCN      #1{\unskip}     \fi
\ifx \shownote     \undefined \def \shownote      #1{#1}          \fi
\ifx \showarticletitle \undefined \def \showarticletitle #1{#1}   \fi
\ifx \showURL      \undefined \def \showURL       {\relax}        \fi
\providecommand\bibfield[2]{#2}
\providecommand\bibinfo[2]{#2}
\providecommand\natexlab[1]{#1}
\providecommand\showeprint[2][]{arXiv:#2}

\bibitem[Akkas and Azad(2024)]%
        {akkas2024gnnshap}
\bibfield{author}{\bibinfo{person}{Selahattin Akkas} {and} \bibinfo{person}{Ariful Azad}.} \bibinfo{year}{2024}\natexlab{}.
\newblock \showarticletitle{Gnnshap: Scalable and accurate gnn explanation using shapley values}. In \bibinfo{booktitle}{\emph{Proceedings of the ACM Web Conference 2024}}. \bibinfo{pages}{827--838}.
\newblock


\bibitem[Cai et~al\mbox{.}(2023)]%
        {cai2023app_gnn_rec3}
\bibfield{author}{\bibinfo{person}{Xuheng Cai}, \bibinfo{person}{Chao Huang}, \bibinfo{person}{Lianghao Xia}, {and} \bibinfo{person}{Xubin Ren}.} \bibinfo{year}{2023}\natexlab{}.
\newblock \showarticletitle{LightGCL: Simple Yet Effective Graph Contrastive Learning for Recommendation}. In \bibinfo{booktitle}{\emph{International Conference on Learning Representations, ICLR}}.
\newblock


\bibitem[Chakraborty et~al\mbox{.}(2018)]%
        {chakraborty2018adversarial}
\bibfield{author}{\bibinfo{person}{Anirban Chakraborty}, \bibinfo{person}{Manaar Alam}, \bibinfo{person}{Vishal Dey}, \bibinfo{person}{Anupam Chattopadhyay}, {and} \bibinfo{person}{Debdeep Mukhopadhyay}.} \bibinfo{year}{2018}\natexlab{}.
\newblock \showarticletitle{Adversarial attacks and defences: A survey}.
\newblock \bibinfo{journal}{\emph{arXiv preprint arXiv:1810.00069}} (\bibinfo{year}{2018}).
\newblock


\bibitem[Chen et~al\mbox{.}(2020)]%
        {chen2020survey}
\bibfield{author}{\bibinfo{person}{Liang Chen}, \bibinfo{person}{Jintang Li}, \bibinfo{person}{Jiaying Peng}, \bibinfo{person}{Tao Xie}, \bibinfo{person}{Zengxu Cao}, \bibinfo{person}{Kun Xu}, \bibinfo{person}{Xiangnan He}, \bibinfo{person}{Zibin Zheng}, {and} \bibinfo{person}{Bingzhe Wu}.} \bibinfo{year}{2020}\natexlab{}.
\newblock \showarticletitle{A survey of adversarial learning on graphs}.
\newblock \bibinfo{journal}{\emph{arXiv preprint arXiv:2003.05730}} (\bibinfo{year}{2020}).
\newblock


\bibitem[Chen et~al\mbox{.}(2024b)]%
        {chen2024llaga}
\bibfield{author}{\bibinfo{person}{Runjin Chen}, \bibinfo{person}{Tong Zhao}, \bibinfo{person}{Ajay Jaiswal}, \bibinfo{person}{Neil Shah}, {and} \bibinfo{person}{Zhangyang Wang}.} \bibinfo{year}{2024}\natexlab{b}.
\newblock \showarticletitle{Llaga: Large language and graph assistant}.
\newblock \bibinfo{journal}{\emph{arXiv preprint arXiv:2402.08170}} (\bibinfo{year}{2024}).
\newblock


\bibitem[Chen et~al\mbox{.}(2022)]%
        {chen2022understanding}
\bibfield{author}{\bibinfo{person}{Yongqiang Chen}, \bibinfo{person}{Han Yang}, \bibinfo{person}{Yonggang Zhang}, \bibinfo{person}{Kaili Ma}, \bibinfo{person}{Tongliang Liu}, \bibinfo{person}{Bo Han}, {and} \bibinfo{person}{James Cheng}.} \bibinfo{year}{2022}\natexlab{}.
\newblock \showarticletitle{Understanding and improving graph injection attack by promoting unnoticeability}.
\newblock \bibinfo{journal}{\emph{arXiv preprint arXiv:2202.08057}} (\bibinfo{year}{2022}).
\newblock


\bibitem[Chen et~al\mbox{.}(2024a)]%
        {chen2024exploring}
\bibfield{author}{\bibinfo{person}{Zhikai Chen}, \bibinfo{person}{Haitao Mao}, \bibinfo{person}{Hang Li}, \bibinfo{person}{Wei Jin}, \bibinfo{person}{Hongzhi Wen}, \bibinfo{person}{Xiaochi Wei}, \bibinfo{person}{Shuaiqiang Wang}, \bibinfo{person}{Dawei Yin}, \bibinfo{person}{Wenqi Fan}, \bibinfo{person}{Hui Liu}, {et~al\mbox{.}}} \bibinfo{year}{2024}\natexlab{a}.
\newblock \showarticletitle{Exploring the potential of large language models (llms) in learning on graphs}.
\newblock \bibinfo{journal}{\emph{ACM SIGKDD Explorations Newsletter}} \bibinfo{volume}{25}, \bibinfo{number}{2} (\bibinfo{year}{2024}), \bibinfo{pages}{42--61}.
\newblock


\bibitem[Chen et~al\mbox{.}(2023)]%
        {chen2023label}
\bibfield{author}{\bibinfo{person}{Zhikai Chen}, \bibinfo{person}{Haitao Mao}, \bibinfo{person}{Hongzhi Wen}, \bibinfo{person}{Haoyu Han}, \bibinfo{person}{Wei Jin}, \bibinfo{person}{Haiyang Zhang}, \bibinfo{person}{Hui Liu}, {and} \bibinfo{person}{Jiliang Tang}.} \bibinfo{year}{2023}\natexlab{}.
\newblock \showarticletitle{Label-free node classification on graphs with large language models (llms)}.
\newblock \bibinfo{journal}{\emph{arXiv preprint arXiv:2310.04668}} (\bibinfo{year}{2023}).
\newblock


\bibitem[Chien et~al\mbox{.}(2021)]%
        {chien2021node}
\bibfield{author}{\bibinfo{person}{Eli Chien}, \bibinfo{person}{Wei-Cheng Chang}, \bibinfo{person}{Cho-Jui Hsieh}, \bibinfo{person}{Hsiang-Fu Yu}, \bibinfo{person}{Jiong Zhang}, \bibinfo{person}{Olgica Milenkovic}, {and} \bibinfo{person}{Inderjit~S Dhillon}.} \bibinfo{year}{2021}\natexlab{}.
\newblock \showarticletitle{Node feature extraction by self-supervised multi-scale neighborhood prediction}.
\newblock \bibinfo{journal}{\emph{arXiv preprint arXiv:2111.00064}} (\bibinfo{year}{2021}).
\newblock


\bibitem[Chien et~al\mbox{.}(2022)]%
        {chien2022node}
\bibfield{author}{\bibinfo{person}{Eli Chien}, \bibinfo{person}{Wei-Cheng Chang}, \bibinfo{person}{Cho-Jui Hsieh}, \bibinfo{person}{Hsiang-Fu Yu}, \bibinfo{person}{Jiong Zhang}, \bibinfo{person}{Olgica Milenkovic}, {and} \bibinfo{person}{Inderjit~S Dhillon}.} \bibinfo{year}{2022}\natexlab{}.
\newblock \showarticletitle{Node Feature Extraction by Self-Supervised Multi-scale Neighborhood Prediction}. In \bibinfo{booktitle}{\emph{International Conference on Learning Representations}}.
\newblock


\bibitem[Dai et~al\mbox{.}(2018)]%
        {dai2018adversarial}
\bibfield{author}{\bibinfo{person}{Hanjun Dai}, \bibinfo{person}{Hui Li}, \bibinfo{person}{Tian Tian}, \bibinfo{person}{Xin Huang}, \bibinfo{person}{Lin Wang}, \bibinfo{person}{Jun Zhu}, {and} \bibinfo{person}{Le Song}.} \bibinfo{year}{2018}\natexlab{}.
\newblock \showarticletitle{Adversarial attack on graph structured data}. In \bibinfo{booktitle}{\emph{International conference on machine learning}}. PMLR, \bibinfo{pages}{1115--1124}.
\newblock


\bibitem[Defferrard et~al\mbox{.}(2016)]%
        {2016Convolutional_ChebNet}
\bibfield{author}{\bibinfo{person}{Michal Defferrard}, \bibinfo{person}{X. Bresson}, {and} \bibinfo{person}{P. Vandergheynst}.} \bibinfo{year}{2016}\natexlab{}.
\newblock \showarticletitle{Convolutional Neural Networks on Graphs with Fast Localized Spectral Filtering}.
\newblock \bibinfo{journal}{\emph{Advances in Neural Information Processing Systems, NeurIPS}} (\bibinfo{year}{2016}).
\newblock


\bibitem[Fout et~al\mbox{.}(2017)]%
        {NIPS2017_f5077839}
\bibfield{author}{\bibinfo{person}{Alex Fout}, \bibinfo{person}{Jonathon Byrd}, \bibinfo{person}{Basir Shariat}, {and} \bibinfo{person}{Asa Ben-Hur}.} \bibinfo{year}{2017}\natexlab{}.
\newblock \showarticletitle{Protein Interface Prediction using Graph Convolutional Networks}. In \bibinfo{booktitle}{\emph{Advances in Neural Information Processing Systems}}, Vol.~\bibinfo{volume}{30}. \bibinfo{publisher}{Curran Associates, Inc.}
\newblock


\bibitem[Garg and Ramakrishnan(2020)]%
        {garg2020bae}
\bibfield{author}{\bibinfo{person}{Siddhant Garg} {and} \bibinfo{person}{Goutham Ramakrishnan}.} \bibinfo{year}{2020}\natexlab{}.
\newblock \showarticletitle{BAE: BERT-based adversarial examples for text classification}.
\newblock \bibinfo{journal}{\emph{arXiv preprint arXiv:2004.01970}} (\bibinfo{year}{2020}).
\newblock


\bibitem[Goodfellow et~al\mbox{.}(2014)]%
        {goodfellow2014explaining}
\bibfield{author}{\bibinfo{person}{Ian~J Goodfellow}, \bibinfo{person}{Jonathon Shlens}, {and} \bibinfo{person}{Christian Szegedy}.} \bibinfo{year}{2014}\natexlab{}.
\newblock \showarticletitle{Explaining and harnessing adversarial examples}.
\newblock \bibinfo{journal}{\emph{arXiv preprint arXiv:1412.6572}} (\bibinfo{year}{2014}).
\newblock


\bibitem[Hamilton et~al\mbox{.}(2017)]%
        {hamilton2017graphsage}
\bibfield{author}{\bibinfo{person}{Will Hamilton}, \bibinfo{person}{Zhitao Ying}, {and} \bibinfo{person}{Jure Leskovec}.} \bibinfo{year}{2017}\natexlab{}.
\newblock \showarticletitle{Inductive representation learning on large graphs}.
\newblock \bibinfo{journal}{\emph{Advances in Neural Information Processing Systems, NeurIPS}} (\bibinfo{year}{2017}).
\newblock


\bibitem[Harris(1954)]%
        {harris1954distributional}
\bibfield{author}{\bibinfo{person}{Zellig~S Harris}.} \bibinfo{year}{1954}\natexlab{}.
\newblock \showarticletitle{Distributional structure}.
\newblock \bibinfo{journal}{\emph{Word}} \bibinfo{volume}{10}, \bibinfo{number}{2-3} (\bibinfo{year}{1954}), \bibinfo{pages}{146--162}.
\newblock


\bibitem[He et~al\mbox{.}(2023)]%
        {he2023harnessing}
\bibfield{author}{\bibinfo{person}{Xiaoxin He}, \bibinfo{person}{Xavier Bresson}, \bibinfo{person}{Thomas Laurent}, \bibinfo{person}{Adam Perold}, \bibinfo{person}{Yann LeCun}, {and} \bibinfo{person}{Bryan Hooi}.} \bibinfo{year}{2023}\natexlab{}.
\newblock \showarticletitle{Harnessing explanations: Llm-to-lm interpreter for enhanced text-attributed graph representation learning}.
\newblock \bibinfo{journal}{\emph{arXiv preprint arXiv:2305.19523}} (\bibinfo{year}{2023}).
\newblock


\bibitem[He et~al\mbox{.}(2020)]%
        {LightGCN}
\bibfield{author}{\bibinfo{person}{Xiangnan He}, \bibinfo{person}{Kuan Deng}, \bibinfo{person}{Xiang Wang}, \bibinfo{person}{Yan Li}, \bibinfo{person}{YongDong Zhang}, {and} \bibinfo{person}{Meng Wang}.} \bibinfo{year}{2020}\natexlab{}.
\newblock \showarticletitle{LightGCN: Simplifying and Powering Graph Convolution Network for Recommendation}. In \bibinfo{booktitle}{\emph{Proceedings of the 43rd International ACM SIGIR Conference on Research and Development in Information Retrieval}}. \bibinfo{publisher}{Association for Computing Machinery}, \bibinfo{pages}{639–648}.
\newblock
\showISBNx{9781450380164}


\bibitem[Hu et~al\mbox{.}(2020)]%
        {hu2020ogb}
\bibfield{author}{\bibinfo{person}{Weihua Hu}, \bibinfo{person}{Matthias Fey}, \bibinfo{person}{Marinka Zitnik}, \bibinfo{person}{Yuxiao Dong}, \bibinfo{person}{Hongyu Ren}, \bibinfo{person}{Bowen Liu}, \bibinfo{person}{Michele Catasta}, {and} \bibinfo{person}{Jure Leskovec}.} \bibinfo{year}{2020}\natexlab{}.
\newblock \showarticletitle{Open graph benchmark: Datasets for machine learning on graphs}.
\newblock \bibinfo{journal}{\emph{Advances in Neural Information Processing Systems, NeurIPS}} (\bibinfo{year}{2020}).
\newblock


\bibitem[Huang et~al\mbox{.}(2020)]%
        {huang2020survey}
\bibfield{author}{\bibinfo{person}{Xiaowei Huang}, \bibinfo{person}{Daniel Kroening}, \bibinfo{person}{Wenjie Ruan}, \bibinfo{person}{James Sharp}, \bibinfo{person}{Youcheng Sun}, \bibinfo{person}{Emese Thamo}, \bibinfo{person}{Min Wu}, {and} \bibinfo{person}{Xinping Yi}.} \bibinfo{year}{2020}\natexlab{}.
\newblock \showarticletitle{A survey of safety and trustworthiness of deep neural networks: Verification, testing, adversarial attack and defence, and interpretability}.
\newblock \bibinfo{journal}{\emph{Computer Science Review}}  \bibinfo{volume}{37} (\bibinfo{year}{2020}), \bibinfo{pages}{100270}.
\newblock


\bibitem[Jin et~al\mbox{.}(2024)]%
        {jin2024large}
\bibfield{author}{\bibinfo{person}{Bowen Jin}, \bibinfo{person}{Gang Liu}, \bibinfo{person}{Chi Han}, \bibinfo{person}{Meng Jiang}, \bibinfo{person}{Heng Ji}, {and} \bibinfo{person}{Jiawei Han}.} \bibinfo{year}{2024}\natexlab{}.
\newblock \showarticletitle{Large language models on graphs: A comprehensive survey}.
\newblock \bibinfo{journal}{\emph{IEEE Transactions on Knowledge and Data Engineering}} (\bibinfo{year}{2024}).
\newblock


\bibitem[Jin et~al\mbox{.}(2020)]%
        {jin2020bert}
\bibfield{author}{\bibinfo{person}{Di Jin}, \bibinfo{person}{Zhijing Jin}, \bibinfo{person}{Joey~Tianyi Zhou}, {and} \bibinfo{person}{Peter Szolovits}.} \bibinfo{year}{2020}\natexlab{}.
\newblock \showarticletitle{Is bert really robust? a strong baseline for natural language attack on text classification and entailment}. In \bibinfo{booktitle}{\emph{Proceedings of the AAAI conference on artificial intelligence}}, Vol.~\bibinfo{volume}{34}. \bibinfo{pages}{8018--8025}.
\newblock


\bibitem[Kipf and Welling(2017)]%
        {kipf2016gcn}
\bibfield{author}{\bibinfo{person}{Thomas~N Kipf} {and} \bibinfo{person}{Max Welling}.} \bibinfo{year}{2017}\natexlab{}.
\newblock \showarticletitle{Semi-supervised classification with graph convolutional networks}. In \bibinfo{booktitle}{\emph{International Conference on Learning Representations, ICLR}}.
\newblock


\bibitem[Kurakin et~al\mbox{.}(2018)]%
        {kurakin2018adversarial}
\bibfield{author}{\bibinfo{person}{Alexey Kurakin}, \bibinfo{person}{Ian~J Goodfellow}, {and} \bibinfo{person}{Samy Bengio}.} \bibinfo{year}{2018}\natexlab{}.
\newblock \showarticletitle{Adversarial examples in the physical world}.
\newblock In \bibinfo{booktitle}{\emph{Artificial intelligence safety and security}}. \bibinfo{publisher}{Chapman and Hall/CRC}, \bibinfo{pages}{99--112}.
\newblock


\bibitem[Lei et~al\mbox{.}(2024)]%
        {lei2024intruding}
\bibfield{author}{\bibinfo{person}{Runlin Lei}, \bibinfo{person}{Yuwei Hu}, \bibinfo{person}{Yuchen Ren}, {and} \bibinfo{person}{Zhewei Wei}.} \bibinfo{year}{2024}\natexlab{}.
\newblock \showarticletitle{Intruding with words: Towards understanding graph injection attacks at the text level}.
\newblock \bibinfo{journal}{\emph{Advances in Neural Information Processing Systems}}  \bibinfo{volume}{37} (\bibinfo{year}{2024}), \bibinfo{pages}{49214--49251}.
\newblock


\bibitem[Li et~al\mbox{.}(2020a)]%
        {li2020bert}
\bibfield{author}{\bibinfo{person}{Linyang Li}, \bibinfo{person}{Ruotian Ma}, \bibinfo{person}{Qipeng Guo}, \bibinfo{person}{Xiangyang Xue}, {and} \bibinfo{person}{Xipeng Qiu}.} \bibinfo{year}{2020}\natexlab{a}.
\newblock \showarticletitle{Bert-attack: Adversarial attack against bert using bert}.
\newblock \bibinfo{journal}{\emph{arXiv preprint arXiv:2004.09984}} (\bibinfo{year}{2020}).
\newblock


\bibitem[Li et~al\mbox{.}(2022)]%
        {li2022distilling}
\bibfield{author}{\bibinfo{person}{Quan Li}, \bibinfo{person}{Xiaoting Li}, \bibinfo{person}{Lingwei Chen}, {and} \bibinfo{person}{Dinghao Wu}.} \bibinfo{year}{2022}\natexlab{}.
\newblock \showarticletitle{Distilling knowledge on text graph for social media attribute inference}. In \bibinfo{booktitle}{\emph{Proceedings of the 45th International ACM SIGIR Conference on Research and Development in Information Retrieval}}. \bibinfo{pages}{2024--2028}.
\newblock


\bibitem[Li et~al\mbox{.}(2020b)]%
        {app10041327}
\bibfield{author}{\bibinfo{person}{Xuefeng Li}, \bibinfo{person}{Yang Xin}, \bibinfo{person}{Chensu Zhao}, \bibinfo{person}{Yixian Yang}, {and} \bibinfo{person}{Yuling Chen}.} \bibinfo{year}{2020}\natexlab{b}.
\newblock \showarticletitle{Graph Convolutional Networks for Privacy Metrics in Online Social Networks}.
\newblock \bibinfo{journal}{\emph{Applied Sciences}} \bibinfo{volume}{10}, \bibinfo{number}{4} (\bibinfo{year}{2020}).
\newblock
\showISSN{2076-3417}


\bibitem[Liu et~al\mbox{.}(2019)]%
        {liu2019roberta}
\bibfield{author}{\bibinfo{person}{Yinhan Liu}, \bibinfo{person}{Myle Ott}, \bibinfo{person}{Naman Goyal}, \bibinfo{person}{Jingfei Du}, \bibinfo{person}{Mandar Joshi}, \bibinfo{person}{Danqi Chen}, \bibinfo{person}{Omer Levy}, \bibinfo{person}{Mike Lewis}, \bibinfo{person}{Luke Zettlemoyer}, {and} \bibinfo{person}{Veselin Stoyanov}.} \bibinfo{year}{2019}\natexlab{}.
\newblock \showarticletitle{Roberta: A robustly optimized bert pretraining approach}.
\newblock \bibinfo{journal}{\emph{arXiv preprint arXiv:1907.11692}} (\bibinfo{year}{2019}).
\newblock


\bibitem[Liu et~al\mbox{.}(2023)]%
        {liu2023graphprompt}
\bibfield{author}{\bibinfo{person}{Zemin Liu}, \bibinfo{person}{Xingtong Yu}, \bibinfo{person}{Yuan Fang}, {and} \bibinfo{person}{Xinming Zhang}.} \bibinfo{year}{2023}\natexlab{}.
\newblock \showarticletitle{Graphprompt: Unifying pre-training and downstream tasks for graph neural networks}. In \bibinfo{booktitle}{\emph{Proceedings of the ACM web conference 2023}}. \bibinfo{pages}{417--428}.
\newblock


\bibitem[Lundberg and Lee(2017a)]%
        {lundberg2017unified}
\bibfield{author}{\bibinfo{person}{Scott~M Lundberg} {and} \bibinfo{person}{Su-In Lee}.} \bibinfo{year}{2017}\natexlab{a}.
\newblock \showarticletitle{A unified approach to interpreting model predictions}.
\newblock \bibinfo{journal}{\emph{Advances in neural information processing systems}}  \bibinfo{volume}{30} (\bibinfo{year}{2017}).
\newblock


\bibitem[Lundberg and Lee(2017b)]%
        {NIPS2017_7062}
\bibfield{author}{\bibinfo{person}{Scott~M Lundberg} {and} \bibinfo{person}{Su-In Lee}.} \bibinfo{year}{2017}\natexlab{b}.
\newblock \showarticletitle{A Unified Approach to Interpreting Model Predictions}.
\newblock \bibinfo{publisher}{Curran Associates, Inc.}
\newblock
\urldef\tempurl%
\url{http://papers.nips.cc/paper/7062-a-unified-approach-to-interpreting-model-predictions.pdf}
\showURL{%
\tempurl}


\bibitem[Luo et~al\mbox{.}(2024)]%
        {luo2024classic}
\bibfield{author}{\bibinfo{person}{Yuankai Luo}, \bibinfo{person}{Lei Shi}, {and} \bibinfo{person}{Xiao-Ming Wu}.} \bibinfo{year}{2024}\natexlab{}.
\newblock \showarticletitle{Classic GNNs are Strong Baselines: Reassessing GNNs for Node Classification}.
\newblock \bibinfo{journal}{\emph{arXiv preprint arXiv:2406.08993}} (\bibinfo{year}{2024}).
\newblock


\bibitem[Ma et~al\mbox{.}(2020)]%
        {ma2020towards}
\bibfield{author}{\bibinfo{person}{Jiaqi Ma}, \bibinfo{person}{Shuangrui Ding}, {and} \bibinfo{person}{Qiaozhu Mei}.} \bibinfo{year}{2020}\natexlab{}.
\newblock \showarticletitle{Towards more practical adversarial attacks on graph neural networks}.
\newblock \bibinfo{journal}{\emph{Advances in neural information processing systems}}  \bibinfo{volume}{33} (\bibinfo{year}{2020}), \bibinfo{pages}{4756--4766}.
\newblock


\bibitem[Mikolov et~al\mbox{.}(2013)]%
        {mikolov2013distributed}
\bibfield{author}{\bibinfo{person}{Tomas Mikolov}, \bibinfo{person}{Ilya Sutskever}, \bibinfo{person}{Kai Chen}, \bibinfo{person}{Greg~S Corrado}, {and} \bibinfo{person}{Jeff Dean}.} \bibinfo{year}{2013}\natexlab{}.
\newblock \showarticletitle{Distributed representations of words and phrases and their compositionality}.
\newblock \bibinfo{journal}{\emph{Advances in neural information processing systems}}  \bibinfo{volume}{26} (\bibinfo{year}{2013}).
\newblock


\bibitem[Mu et~al\mbox{.}(2021)]%
        {mu2021hard}
\bibfield{author}{\bibinfo{person}{Jiaming Mu}, \bibinfo{person}{Binghui Wang}, \bibinfo{person}{Qi Li}, \bibinfo{person}{Kun Sun}, \bibinfo{person}{Mingwei Xu}, {and} \bibinfo{person}{Zhuotao Liu}.} \bibinfo{year}{2021}\natexlab{}.
\newblock \showarticletitle{A hard label black-box adversarial attack against graph neural networks}. In \bibinfo{booktitle}{\emph{Proceedings of the 2021 ACM SIGSAC Conference on Computer and Communications Security}}. \bibinfo{pages}{108--125}.
\newblock


\bibitem[Murdoch et~al\mbox{.}(2019)]%
        {murdoch2019definitions}
\bibfield{author}{\bibinfo{person}{W~James Murdoch}, \bibinfo{person}{Chandan Singh}, \bibinfo{person}{Karl Kumbier}, \bibinfo{person}{Reza Abbasi-Asl}, {and} \bibinfo{person}{Bin Yu}.} \bibinfo{year}{2019}\natexlab{}.
\newblock \showarticletitle{Definitions, methods, and applications in interpretable machine learning}.
\newblock \bibinfo{journal}{\emph{Proceedings of the National Academy of Sciences}} \bibinfo{volume}{116}, \bibinfo{number}{44} (\bibinfo{year}{2019}), \bibinfo{pages}{22071--22080}.
\newblock


\bibitem[Muschalik et~al\mbox{.}(2025)]%
        {muschalik2025exact}
\bibfield{author}{\bibinfo{person}{Maximilian Muschalik}, \bibinfo{person}{Fabian Fumagalli}, \bibinfo{person}{Paolo Frazzetto}, \bibinfo{person}{Janine Strotherm}, \bibinfo{person}{Luca Hermes}, \bibinfo{person}{Alessandro Sperduti}, \bibinfo{person}{Eyke H{\"u}llermeier}, {and} \bibinfo{person}{Barbara Hammer}.} \bibinfo{year}{2025}\natexlab{}.
\newblock \showarticletitle{Exact Computation of Any-Order Shapley Interactions for Graph Neural Networks}.
\newblock \bibinfo{journal}{\emph{arXiv preprint arXiv:2501.16944}} (\bibinfo{year}{2025}).
\newblock


\bibitem[Papernot et~al\mbox{.}(2017)]%
        {papernot2017practical}
\bibfield{author}{\bibinfo{person}{Nicolas Papernot}, \bibinfo{person}{Patrick McDaniel}, \bibinfo{person}{Ian Goodfellow}, \bibinfo{person}{Somesh Jha}, \bibinfo{person}{Z~Berkay Celik}, {and} \bibinfo{person}{Ananthram Swami}.} \bibinfo{year}{2017}\natexlab{}.
\newblock \showarticletitle{Practical black-box attacks against machine learning}. In \bibinfo{booktitle}{\emph{Proceedings of the 2017 ACM on Asia conference on computer and communications security}}. \bibinfo{pages}{506--519}.
\newblock


\bibitem[Qin et~al\mbox{.}(2023)]%
        {qin2023disentangled}
\bibfield{author}{\bibinfo{person}{Yijian Qin}, \bibinfo{person}{Xin Wang}, \bibinfo{person}{Ziwei Zhang}, {and} \bibinfo{person}{Wenwu Zhu}.} \bibinfo{year}{2023}\natexlab{}.
\newblock \showarticletitle{Disentangled representation learning with large language models for text-attributed graphs}.
\newblock \bibinfo{journal}{\emph{arXiv preprint arXiv:2310.18152}} (\bibinfo{year}{2023}).
\newblock


\bibitem[Qu et~al\mbox{.}(2023)]%
        {qu2023app_gnn_bio2}
\bibfield{author}{\bibinfo{person}{Zongshuai Qu}, \bibinfo{person}{Tao Yao}, \bibinfo{person}{Xinghui Liu}, {and} \bibinfo{person}{Gang Wang}.} \bibinfo{year}{2023}\natexlab{}.
\newblock \showarticletitle{A Graph Convolutional Network Based on Univariate Neurodegeneration Biomarker for Alzheimer’s Disease Diagnosis}.
\newblock \bibinfo{journal}{\emph{IEEE Journal of Translational Engineering in Health and Medicine}} (\bibinfo{year}{2023}).
\newblock


\bibitem[Radford et~al\mbox{.}(2019)]%
        {radford2019language}
\bibfield{author}{\bibinfo{person}{Alec Radford}, \bibinfo{person}{Jeffrey Wu}, \bibinfo{person}{Rewon Child}, \bibinfo{person}{David Luan}, \bibinfo{person}{Dario Amodei}, \bibinfo{person}{Ilya Sutskever}, {et~al\mbox{.}}} \bibinfo{year}{2019}\natexlab{}.
\newblock \showarticletitle{Language models are unsupervised multitask learners}.
\newblock \bibinfo{journal}{\emph{OpenAI blog}} \bibinfo{volume}{1}, \bibinfo{number}{8} (\bibinfo{year}{2019}), \bibinfo{pages}{9}.
\newblock


\bibitem[Reimers and Gurevych(2019)]%
        {reimers-gurevych-2019-sentence}
\bibfield{author}{\bibinfo{person}{Nils Reimers} {and} \bibinfo{person}{Iryna Gurevych}.} \bibinfo{year}{2019}\natexlab{}.
\newblock \showarticletitle{Sentence-{BERT}: Sentence Embeddings using {S}iamese {BERT}-Networks}. In \bibinfo{booktitle}{\emph{Proceedings of the 2019 Conference on Empirical Methods in Natural Language Processing and the 9th International Joint Conference on Natural Language Processing (EMNLP-IJCNLP)}}, \bibfield{editor}{\bibinfo{person}{Kentaro Inui}, \bibinfo{person}{Jing Jiang}, \bibinfo{person}{Vincent Ng}, {and} \bibinfo{person}{Xiaojun Wan}} (Eds.). \bibinfo{address}{Hong Kong, China}, \bibinfo{pages}{3982--3992}.
\newblock
\href{https://doi.org/10.18653/v1/D19-1410}{doi:\nolinkurl{10.18653/v1/D19-1410}}


\bibitem[Ribeiro et~al\mbox{.}(2016)]%
        {ribeiro2016should}
\bibfield{author}{\bibinfo{person}{Marco~Tulio Ribeiro}, \bibinfo{person}{Sameer Singh}, {and} \bibinfo{person}{Carlos Guestrin}.} \bibinfo{year}{2016}\natexlab{}.
\newblock \showarticletitle{" Why should i trust you?" Explaining the predictions of any classifier}. In \bibinfo{booktitle}{\emph{Proceedings of the 22nd ACM SIGKDD international conference on knowledge discovery and data mining}}. \bibinfo{pages}{1135--1144}.
\newblock


\bibitem[Schlichtkrull et~al\mbox{.}(2018)]%
        {schlichtkrull2018modeling}
\bibfield{author}{\bibinfo{person}{Michael Schlichtkrull}, \bibinfo{person}{Thomas~N Kipf}, \bibinfo{person}{Peter Bloem}, \bibinfo{person}{Rianne Van Den~Berg}, \bibinfo{person}{Ivan Titov}, {and} \bibinfo{person}{Max Welling}.} \bibinfo{year}{2018}\natexlab{}.
\newblock \showarticletitle{Modeling relational data with graph convolutional networks}. In \bibinfo{booktitle}{\emph{European semantic web conference}}. Springer, \bibinfo{pages}{593--607}.
\newblock


\bibitem[Sen et~al\mbox{.}(2008)]%
        {sen2008collective}
\bibfield{author}{\bibinfo{person}{Prithviraj Sen}, \bibinfo{person}{Galileo Namata}, \bibinfo{person}{Mustafa Bilgic}, \bibinfo{person}{Lise Getoor}, \bibinfo{person}{Brian Galligher}, {and} \bibinfo{person}{Tina Eliassi-Rad}.} \bibinfo{year}{2008}\natexlab{}.
\newblock \showarticletitle{Collective classification in network data}.
\newblock \bibinfo{journal}{\emph{AI magazine}} \bibinfo{volume}{29}, \bibinfo{number}{3} (\bibinfo{year}{2008}), \bibinfo{pages}{93--93}.
\newblock


\bibitem[Shapley et~al\mbox{.}(1953)]%
        {shapley1953value}
\bibfield{author}{\bibinfo{person}{Lloyd~S Shapley} {et~al\mbox{.}}} \bibinfo{year}{1953}\natexlab{}.
\newblock \showarticletitle{A value for n-person games}.
\newblock  (\bibinfo{year}{1953}).
\newblock


\bibitem[Sharma(2018)]%
        {sharma2018gradient}
\bibfield{author}{\bibinfo{person}{Yash Sharma}.} \bibinfo{year}{2018}\natexlab{}.
\newblock \showarticletitle{Gradient-based Adversarial Attacks to Deep Neural Networks in Limited Access Settings}.
\newblock \bibinfo{journal}{\emph{THE COOPER UNION ALBERT NERKEN SCHOOL OF ENGINEERING}} (\bibinfo{year}{2018}).
\newblock


\bibitem[{\v{S}}trumbelj and Kononenko(2014)]%
        {vstrumbelj2014explaining}
\bibfield{author}{\bibinfo{person}{Erik {\v{S}}trumbelj} {and} \bibinfo{person}{Igor Kononenko}.} \bibinfo{year}{2014}\natexlab{}.
\newblock \showarticletitle{Explaining prediction models and individual predictions with feature contributions}.
\newblock \bibinfo{journal}{\emph{Knowledge and information systems}} \bibinfo{volume}{41}, \bibinfo{number}{3} (\bibinfo{year}{2014}), \bibinfo{pages}{647--665}.
\newblock


\bibitem[Su et~al\mbox{.}(2024)]%
        {su2024dcl}
\bibfield{author}{\bibinfo{person}{Daohan Su}, \bibinfo{person}{Bowen Fan}, \bibinfo{person}{Zhi Zhang}, \bibinfo{person}{Haoyan Fu}, {and} \bibinfo{person}{Zhida Qin}.} \bibinfo{year}{2024}\natexlab{}.
\newblock \showarticletitle{DCL: Diversified Graph Recommendation With Contrastive Learning}.
\newblock \bibinfo{journal}{\emph{IEEE Transactions on Computational Social Systems}} (\bibinfo{year}{2024}).
\newblock


\bibitem[Sun et~al\mbox{.}(2023)]%
        {sun2023adpa}
\bibfield{author}{\bibinfo{person}{Henan Sun}, \bibinfo{person}{Xunkai Li}, \bibinfo{person}{Zhengyu Wu}, \bibinfo{person}{Daohan Su}, \bibinfo{person}{Rong-Hua Li}, {and} \bibinfo{person}{Guoren Wang}.} \bibinfo{year}{2023}\natexlab{}.
\newblock \showarticletitle{Breaking the Entanglement of Homophily and Heterophily in Semi-supervised Node Classification}.
\newblock \bibinfo{journal}{\emph{arXiv preprint arXiv:2312.04111}} (\bibinfo{year}{2023}).
\newblock


\bibitem[Sun et~al\mbox{.}(2022)]%
        {sun2022adversarial}
\bibfield{author}{\bibinfo{person}{Lichao Sun}, \bibinfo{person}{Yingtong Dou}, \bibinfo{person}{Carl Yang}, \bibinfo{person}{Kai Zhang}, \bibinfo{person}{Ji Wang}, \bibinfo{person}{Philip~S Yu}, \bibinfo{person}{Lifang He}, {and} \bibinfo{person}{Bo Li}.} \bibinfo{year}{2022}\natexlab{}.
\newblock \showarticletitle{Adversarial attack and defense on graph data: A survey}.
\newblock \bibinfo{journal}{\emph{IEEE Transactions on Knowledge and Data Engineering}} \bibinfo{volume}{35}, \bibinfo{number}{8} (\bibinfo{year}{2022}), \bibinfo{pages}{7693--7711}.
\newblock


\bibitem[Sun et~al\mbox{.}(2020)]%
        {sun2020adversarial}
\bibfield{author}{\bibinfo{person}{Yiwei Sun}, \bibinfo{person}{Suhang Wang}, \bibinfo{person}{Xianfeng Tang}, \bibinfo{person}{Tsung-Yu Hsieh}, {and} \bibinfo{person}{Vasant Honavar}.} \bibinfo{year}{2020}\natexlab{}.
\newblock \showarticletitle{Adversarial attacks on graph neural networks via node injections: A hierarchical reinforcement learning approach}. In \bibinfo{booktitle}{\emph{Proceedings of the Web Conference 2020}}. \bibinfo{pages}{673--683}.
\newblock


\bibitem[Veli{\v{c}}kovi{\'c} et~al\mbox{.}(2018)]%
        {velivckovic2017gat}
\bibfield{author}{\bibinfo{person}{Petar Veli{\v{c}}kovi{\'c}}, \bibinfo{person}{Guillem Cucurull}, \bibinfo{person}{Arantxa Casanova}, \bibinfo{person}{Adriana Romero}, \bibinfo{person}{Pietro Lio}, {and} \bibinfo{person}{Yoshua Bengio}.} \bibinfo{year}{2018}\natexlab{}.
\newblock \showarticletitle{Graph attention networks}. In \bibinfo{booktitle}{\emph{International Conference on Learning Representations, ICLR}}.
\newblock


\bibitem[Wang et~al\mbox{.}(2022)]%
        {wang2022semattack}
\bibfield{author}{\bibinfo{person}{Boxin Wang}, \bibinfo{person}{Chejian Xu}, \bibinfo{person}{Xiangyu Liu}, \bibinfo{person}{Yu Cheng}, {and} \bibinfo{person}{Bo Li}.} \bibinfo{year}{2022}\natexlab{}.
\newblock \showarticletitle{SemAttack: Natural textual attacks via different semantic spaces}.
\newblock \bibinfo{journal}{\emph{arXiv preprint arXiv:2205.01287}} (\bibinfo{year}{2022}).
\newblock


\bibitem[Wang et~al\mbox{.}(2020a)]%
        {wang2020scalable}
\bibfield{author}{\bibinfo{person}{Jihong Wang}, \bibinfo{person}{Minnan Luo}, \bibinfo{person}{Fnu Suya}, \bibinfo{person}{Jundong Li}, \bibinfo{person}{Zijiang Yang}, {and} \bibinfo{person}{Qinghua Zheng}.} \bibinfo{year}{2020}\natexlab{a}.
\newblock \showarticletitle{Scalable attack on graph data by injecting vicious nodes}.
\newblock \bibinfo{journal}{\emph{Data Mining and Knowledge Discovery}} \bibinfo{volume}{34}, \bibinfo{number}{5} (\bibinfo{year}{2020}), \bibinfo{pages}{1363--1389}.
\newblock


\bibitem[Wang et~al\mbox{.}(2020b)]%
        {wang2020microsoft}
\bibfield{author}{\bibinfo{person}{Kuansan Wang}, \bibinfo{person}{Zhihong Shen}, \bibinfo{person}{Chiyuan Huang}, \bibinfo{person}{Chieh-Han Wu}, \bibinfo{person}{Yuxiao Dong}, {and} \bibinfo{person}{Anshul Kanakia}.} \bibinfo{year}{2020}\natexlab{b}.
\newblock \showarticletitle{Microsoft academic graph: When experts are not enough}.
\newblock \bibinfo{journal}{\emph{Quantitative Science Studies}} \bibinfo{volume}{1}, \bibinfo{number}{1} (\bibinfo{year}{2020}), \bibinfo{pages}{396--413}.
\newblock


\bibitem[Wang et~al\mbox{.}(2018)]%
        {ijcai2018p142}
\bibfield{author}{\bibinfo{person}{Zhouxia Wang}, \bibinfo{person}{Tianshui Chen}, \bibinfo{person}{Jimmy Ren}, \bibinfo{person}{Weihao Yu}, \bibinfo{person}{Hui Cheng}, {and} \bibinfo{person}{Liang Lin}.} \bibinfo{year}{2018}\natexlab{}.
\newblock \showarticletitle{Deep Reasoning with Knowledge Graph for Social Relationship Understanding}. In \bibinfo{booktitle}{\emph{Proceedings of the Twenty-Seventh International Joint Conference on Artificial Intelligence, {IJCAI-18}}}. \bibinfo{publisher}{International Joint Conferences on Artificial Intelligence Organization}, \bibinfo{pages}{1021--1028}.
\newblock


\bibitem[Wu et~al\mbox{.}(2021)]%
        {graphsurvey}
\bibfield{author}{\bibinfo{person}{Zonghan Wu}, \bibinfo{person}{Shirui Pan}, \bibinfo{person}{Fengwen Chen}, \bibinfo{person}{Guodong Long}, \bibinfo{person}{Chengqi Zhang}, {and} \bibinfo{person}{Philip~S. Yu}.} \bibinfo{year}{2021}\natexlab{}.
\newblock \showarticletitle{A Comprehensive Survey on Graph Neural Networks}.
\newblock \bibinfo{journal}{\emph{IEEE Transactions on Neural Networks and Learning Systems}} \bibinfo{volume}{32}, \bibinfo{number}{1} (\bibinfo{year}{2021}), \bibinfo{pages}{4--24}.
\newblock
\href{https://doi.org/10.1109/TNNLS.2020.2978386}{doi:\nolinkurl{10.1109/TNNLS.2020.2978386}}


\bibitem[Xia et~al\mbox{.}(2021)]%
        {Xia_2021}
\bibfield{author}{\bibinfo{person}{Feng Xia}, \bibinfo{person}{Ke Sun}, \bibinfo{person}{Shuo Yu}, \bibinfo{person}{Abdul Aziz}, \bibinfo{person}{Liangtian Wan}, \bibinfo{person}{Shirui Pan}, {and} \bibinfo{person}{Huan Liu}.} \bibinfo{year}{2021}\natexlab{}.
\newblock \showarticletitle{Graph Learning: A Survey}.
\newblock \bibinfo{journal}{\emph{IEEE Transactions on Artificial Intelligence}} \bibinfo{volume}{2}, \bibinfo{number}{2} (\bibinfo{date}{April} \bibinfo{year}{2021}), \bibinfo{pages}{109–127}.
\newblock
\showISSN{2691-4581}
\href{https://doi.org/10.1109/tai.2021.3076021}{doi:\nolinkurl{10.1109/tai.2021.3076021}}


\bibitem[Xu et~al\mbox{.}(2019a)]%
        {xu2019topology}
\bibfield{author}{\bibinfo{person}{Kaidi Xu}, \bibinfo{person}{Hongge Chen}, \bibinfo{person}{Sijia Liu}, \bibinfo{person}{Pin-Yu Chen}, \bibinfo{person}{Tsui-Wei Weng}, \bibinfo{person}{Mingyi Hong}, {and} \bibinfo{person}{Xue Lin}.} \bibinfo{year}{2019}\natexlab{a}.
\newblock \showarticletitle{Topology attack and defense for graph neural networks: An optimization perspective}.
\newblock \bibinfo{journal}{\emph{arXiv preprint arXiv:1906.04214}} (\bibinfo{year}{2019}).
\newblock


\bibitem[Xu et~al\mbox{.}(2019b)]%
        {xu2018gin}
\bibfield{author}{\bibinfo{person}{Keyulu Xu}, \bibinfo{person}{Weihua Hu}, \bibinfo{person}{Jure Leskovec}, {and} \bibinfo{person}{Stefanie Jegelka}.} \bibinfo{year}{2019}\natexlab{b}.
\newblock \showarticletitle{How powerful are graph neural networks?} \bibinfo{publisher}{International Conference on Learning Representations, ICLR}.
\newblock


\bibitem[Yang et~al\mbox{.}(2016)]%
        {Yang16cora}
\bibfield{author}{\bibinfo{person}{Zhilin Yang}, \bibinfo{person}{William~W. Cohen}, {and} \bibinfo{person}{Ruslan Salakhutdinov}.} \bibinfo{year}{2016}\natexlab{}.
\newblock \showarticletitle{Revisiting Semi-Supervised Learning with Graph Embeddings}. In \bibinfo{booktitle}{\emph{International Conference on Machine Learning, ICML}}.
\newblock


\bibitem[Ye et~al\mbox{.}(2022)]%
        {ye2022texthoaxer}
\bibfield{author}{\bibinfo{person}{Muchao Ye}, \bibinfo{person}{Chenglin Miao}, \bibinfo{person}{Ting Wang}, {and} \bibinfo{person}{Fenglong Ma}.} \bibinfo{year}{2022}\natexlab{}.
\newblock \showarticletitle{Texthoaxer: Budgeted hard-label adversarial attacks on text}. In \bibinfo{booktitle}{\emph{Proceedings of the AAAI Conference on Artificial Intelligence}}, Vol.~\bibinfo{volume}{36}. \bibinfo{pages}{3877--3884}.
\newblock


\bibitem[Zhang and Zitnik(2020)]%
        {zhang2020gnnguard}
\bibfield{author}{\bibinfo{person}{Xiang Zhang} {and} \bibinfo{person}{Marinka Zitnik}.} \bibinfo{year}{2020}\natexlab{}.
\newblock \showarticletitle{Gnnguard: Defending graph neural networks against adversarial attacks}.
\newblock \bibinfo{journal}{\emph{Advances in neural information processing systems}}  \bibinfo{volume}{33} (\bibinfo{year}{2020}), \bibinfo{pages}{9263--9275}.
\newblock


\bibitem[Zhao et~al\mbox{.}(2023)]%
        {zhao2023learning}
\bibfield{author}{\bibinfo{person}{Jianan Zhao}, \bibinfo{person}{Meng Qu}, \bibinfo{person}{Chaozhuo Li}, \bibinfo{person}{Hao Yan}, \bibinfo{person}{Qian Liu}, \bibinfo{person}{Rui Li}, \bibinfo{person}{Xing Xie}, {and} \bibinfo{person}{Jian Tang}.} \bibinfo{year}{2023}\natexlab{}.
\newblock \showarticletitle{Learning on Large-scale Text-attributed Graphs via Variational Inference}. In \bibinfo{booktitle}{\emph{The Eleventh International Conference on Learning Representations}}.
\newblock


\bibitem[Zheng et~al\mbox{.}(2021)]%
        {zheng2021graph}
\bibfield{author}{\bibinfo{person}{Qinkai Zheng}, \bibinfo{person}{Xu Zou}, \bibinfo{person}{Yuxiao Dong}, \bibinfo{person}{Yukuo Cen}, \bibinfo{person}{Da Yin}, \bibinfo{person}{Jiarong Xu}, \bibinfo{person}{Yang Yang}, {and} \bibinfo{person}{Jie Tang}.} \bibinfo{year}{2021}\natexlab{}.
\newblock \showarticletitle{Graph robustness benchmark: Benchmarking the adversarial robustness of graph machine learning}.
\newblock \bibinfo{journal}{\emph{arXiv preprint arXiv:2111.04314}} (\bibinfo{year}{2021}).
\newblock


\bibitem[Zhu et~al\mbox{.}(2024)]%
        {zhu2024efficient}
\bibfield{author}{\bibinfo{person}{Yun Zhu}, \bibinfo{person}{Yaoke Wang}, \bibinfo{person}{Haizhou Shi}, {and} \bibinfo{person}{Siliang Tang}.} \bibinfo{year}{2024}\natexlab{}.
\newblock \showarticletitle{Efficient tuning and inference for large language models on textual graphs}.
\newblock \bibinfo{journal}{\emph{arXiv preprint arXiv:2401.15569}} (\bibinfo{year}{2024}).
\newblock


\bibitem[Zitnik et~al\mbox{.}(2018)]%
        {10.1093/bioinformatics/bty294}
\bibfield{author}{\bibinfo{person}{Marinka Zitnik}, \bibinfo{person}{Monica Agrawal}, {and} \bibinfo{person}{Jure Leskovec}.} \bibinfo{year}{2018}\natexlab{}.
\newblock \showarticletitle{{Modeling polypharmacy side effects with graph convolutional networks}}.
\newblock \bibinfo{journal}{\emph{Bioinformatics}} \bibinfo{volume}{34}, \bibinfo{number}{13} (\bibinfo{date}{06} \bibinfo{year}{2018}), \bibinfo{pages}{i457--i466}.
\newblock
\showISSN{1367-4803}


\bibitem[Zou et~al\mbox{.}(2021)]%
        {zou2021tdgia}
\bibfield{author}{\bibinfo{person}{Xu Zou}, \bibinfo{person}{Qinkai Zheng}, \bibinfo{person}{Yuxiao Dong}, \bibinfo{person}{Xinyu Guan}, \bibinfo{person}{Evgeny Kharlamov}, \bibinfo{person}{Jialiang Lu}, {and} \bibinfo{person}{Jie Tang}.} \bibinfo{year}{2021}\natexlab{}.
\newblock \showarticletitle{Tdgia: Effective injection attacks on graph neural networks}. In \bibinfo{booktitle}{\emph{Proceedings of the 27th ACM SIGKDD Conference on Knowledge Discovery \& Data Mining}}. \bibinfo{pages}{2461--2471}.
\newblock


\bibitem[Z{\"u}gner and G{\"u}nnemann(2019)]%
        {zugner_adversarial_2019_metaattck}
\bibfield{author}{\bibinfo{person}{Daniel Z{\"u}gner} {and} \bibinfo{person}{Stephan G{\"u}nnemann}.} \bibinfo{year}{2019}\natexlab{}.
\newblock \showarticletitle{Adversarial Attacks on Graph Neural Networks via Meta Learning}. In \bibinfo{booktitle}{\emph{International Conference on Learning Representations, ICLR}}.
\newblock


\end{thebibliography}

\appendix

\appendix
\clearpage
\section{Datasets} 
\label{app: datasets}
Experiments are carried out on several widely used, representative text-attributed graph datasets (Table~\ref{tab:dataset_stats}).
In these datasets, each node is associated with a short textual description (e.g., paper title or abstract, product description), and edges encode citations.
Node features are obtained by encoding node texts with our encoding model.
We adopt stratified splits of 10\%/10\%/80\% (train/val/test) for Cora, Citeseer, and PubMed, and 20\%/20\%/60\% for ogbn-arxiv.

\begin{table}[h]
  \setlength{\abovecaptionskip}{0.2cm}
  \setlength{\belowcaptionskip}{-0.2cm}
  \caption{Dataset statistics.}
  \label{tab:dataset_stats}
  \setlength{\tabcolsep}{6pt}
  \begin{tabular}{lcccc}
    \arrayrulecolor{wwwgreen}\toprule[0.16em]
    \textbf{Dataset} & \textbf{\#Nodes} & \textbf{\#Edges} & \textbf{\#Classes} & \textbf{Avg. Degree} \\
    \arrayrulecolor{wwwgreen}\midrule[0.16em]
    Cora       & 2,708   & 10,556    & 7  & 3.90 \\
    Citeseer   & 3,186   & 8,450     & 6  & 2.65 \\
    PubMed     & 19,717  & 88,648    & 3  & 4.50 \\
    ogbn-arxiv & 169,343 & 2,315,598 & 40 & 6.89 \\
    \arrayrulecolor{wwwgreen}\bottomrule[0.16em]
  \end{tabular}
  \vspace{-0.3cm}
\end{table}





\section{Attack Methods}
\label{app:baseline}

Two families of baselines are considered on text-attributed graphs: text-side attacks that edit node texts and feature-side attacks that perturb continuous text-derived features. For fair and stealthy comparison, all methods attack the same pre-specified nodes under an untargeted setting with matched cosine-similarity thresholds.

\textbf{HLBB}~\cite{mu2021hard} is a hard-label, decision-based black-box attack that observes only the predicted label and uses a population-based word-substitution search to push the example across the decision boundary while preserving semantic similarity.

\textbf{TextHoaxer}~\cite{ye2022texthoaxer} is a budget-aware hard-label attack that casts discrete word substitution as a continuous optimization in embedding space. It iteratively refines a single candidate with a loss combining semantic similarity, pairwise perturbation, and sparsity to reduce queries while maintaining fluency.

\textbf{SemAttack}~\cite{wang2022semattack} is a semantics-preserving attack that defines perturbations across multiple semantic spaces (typos, WordNet synonyms, and contextualized BERT neighborhoods) and optimizes within these spaces while controlling the magnitude of changes.

\textbf{PA-F}~\cite{ma2020towards} is a black-box baseline that perturbs node features while keeping the graph fixed. The original method includes RWCS-based node selection with a GC-RWCS variant. In our setting, we disable node selection and attack a fixed node set, applying only feature perturbation. Perturbations are further bounded by a cosine-similarity threshold for fairness.

\textbf{FGSM}~\cite{goodfellow2014explaining} is a one-step gradient-sign attack in the continuous feature space. Given features $x$, it computes $\eta=\epsilon\,\mathrm{sign}(\nabla_x J)$ and projects $x'=\Pi_{\mathcal{C}}(x+\eta)$ onto a feasible set that enforces either an $\ell_\infty$ or an $\ell_2$ bound together with a cosine-similarity threshold. The graph topology and raw tokens remain fixed.

\section{Experimental Settings} 

\label{app: exp settings}
In this section, we present a unified description of the experimental setups for each problem to ensure reproducibility and consistency. To guarantee the reliability of the results, each experiment is repeated five times independently, and both the mean and standard deviation are reported. For Q1, we select 10\% of nodes from Cora and Citeseer as target nodes, while for the larger datasets PubMed and ogbn-arxiv, we sample 500 nodes. Regarding the key parameters, the text perturbation ratio and the candidate set size are fixed at 30\% and top-30, respectively. It is worth noting that since PA-F and FGSM perturb the original features, we constrain them using the same similarity measure as IMDGA. For the remaining experiments, to improve efficiency, the number of target nodes is fixed at 100, while all other hyperparameters remain unchanged.
Unless otherwise specified, we choose a 2-layer GCN as the backbone.
\section{Hyperparameter Settings}
Table~\ref{tab:hyperspace} details the hyperparameters and search ranges used in all reported experiments, with notation following the main text.

\label{app:hypersetting}
\begin{table}[h] 
\setlength{\abovecaptionskip}{0.2cm}
\setlength{\belowcaptionskip}{-0.2cm}
\caption{ Search Space for IMDGA.
}
\label{tab:hyperspace}
\begin{tabular}{ccc}
\arrayrulecolor{wwwgreen}\toprule[0.16em]
{Hyperparameter} & Description  & Search Space \\
\arrayrulecolor{wwwgreen} \midrule[0.16em]
$\beta$                                   
& Text Modification Ratio & \{0,0.05,...,0.4\}\\
$\alpha$                        
& Label Flip Weight& \{0,1,...,5\} \\
 top-$k_1$                     
& Candidate Word Number & \{0,5,...,40\} \\
top-$k_2$                   
& Edge Pruning Number & \{0,2,4,6\} \\

$\alpha_1,\alpha_2,\alpha_3$ 
& Scoring Function Weight& Grid Search \\
\arrayrulecolor{wwwgreen} \midrule[0.16em]
\end{tabular}
\vspace{-0.3cm}
\end{table}

\section{Experiment}
Table~\ref{tab:arxiv} reports additional ASR results on ogbn-arxiv under identical budgets and cosine-similarity constraints. \textbf{OOT}: out of time (12-h wall-clock limit exceeded).

\begin{table}[h]
\caption{ASR comparison on ogbn-arxiv.}
\vspace{1mm}
\centering
\fontsize{10pt}{11pt}\selectfont
\renewcommand{\arraystretch}{1.25}
\resizebox{0.48\textwidth}{!}{
\begin{tabular}{llccccc}
\arrayrulecolor{wwwgreen}\toprule[0.16em]
{\textbf{Dataset}} & {\textbf{Methods}} & {\textbf{SBERT}} & {\textbf{BERT}} &{\textbf{RoBERTa}} &{\textbf{DeBERTa}} & {\textbf{DistilBERT}}\\ 
\arrayrulecolor{wwwgreen2}\midrule[0.08em] \addlinespace[-1.7pt]
\arrayrulecolor{wwwgreen3} \midrule[0.08em]
\multirow{7}{*}{\textbf{Arxiv}}
& HLBB 
& {35.12{\scriptsize \(\pm\)0.32}} 
& {21.76{\scriptsize \(\pm\)0.84}} 
& {18.14{\scriptsize \(\pm\)1.57}} 
& {39.76{\scriptsize \(\pm\)0.49}} 
& {37.16{\scriptsize \(\pm\)1.22}} 

 \\
& TextHoaxer 
& {36.41{\scriptsize \(\pm\)1.15}}
& {22.24{\scriptsize \(\pm\)0.82}}
& {22.39{\scriptsize \(\pm\)0.43}}
& {40.03{\scriptsize \(\pm\)0.11}}
& {38.83{\scriptsize \(\pm\)1.56}}
 \\
& SemAttack
& {OOT}
& {OOT}
& {OOT}
& {OOT}
& {OOT}
\\
& FGSM
& {30.44 {\scriptsize \(\pm\)0.26}}
& {35.09{\scriptsize \(\pm\)0.88}}
& {18.81{\scriptsize \(\pm\)1.07}}
& {31.87{\scriptsize \(\pm\)0.70}}
& {30.55{\scriptsize \(\pm\)1.92}}
 \\

& PA-F 
& {4.37 {\scriptsize \(\pm\)0.42}}
& {21.13{\scriptsize \(\pm\)1.16}}
& {15.27{\scriptsize \(\pm\)0.51}}
& {4.56{\scriptsize \(\pm\)0.67}}
& {5.62{\scriptsize \(\pm\)0.45}}
 \\
& IMDGA 
& {\textbf{47.82{\scriptsize \(\pm\)1.03}}}
& {\textbf{45.56{\scriptsize \(\pm\)0.22}}}
& {\textbf{47.32{\scriptsize \(\pm\)0.48}}}
& {\textbf{44.73{\scriptsize \(\pm\)0.11}}}
&{\textbf{48.12{\scriptsize \(\pm\)0.45}}}
\\

\arrayrulecolor{wwwgreen}\bottomrule[0.16em]
\end{tabular}
}
\label{tab:arxiv}
\end{table}

\section{Human Evaluation}
\label{app:human}

We recruit three trained student annotators to evaluate text-side adversarial examples after a short warm-up. 
For each case, sentences are shown in random, blind order, and each annotator rates them independently without discussion. 
Scores are averaged per sentence across annotators and then combined into per-method means across all cases for reporting.

Below are the annotation instructions and the concise guideline for language-quality ratings used herein.

Please rate overall language quality on a 1–5 scale (coherence, fluency, grammar), considering clarity and readability.

\begin{itemize}[leftmargin=1em,itemsep=3pt,topsep=2pt,parsep=0pt]
  \item \textbf{5} — Natural, coherent; no errors; fully fluent.
  \item \textbf{4} — Minor issues; easy to read; few typos.
  \item \textbf{3} — Clear meaning; some roughness; occasional errors.
  \item \textbf{2} — Awkward; hinders understanding; frequent errors.
  \item \textbf{1} — Incoherent; severely ungrammatical; unreadable.
\end{itemize}





\section{Adversarial Examples}\vspace{-6pt}
\begin{table}[H]
\setlength{\tabcolsep}{6pt}
\renewcommand{\arraystretch}{1.12}
\caption{Examples of successful IMDGA attacks on Cora: original tokens in \textbf{bold} and adversarial substitutions in \textcolor{red}{red}.}
\label{tab:adv_examples_textonly}
\begin{tabularx}{\linewidth}{M{0.09\linewidth}!{\vrule width 0.4pt} J }
\toprule
\textbf{Type} & \multicolumn{1}{c}{\textbf{Text}} \\
\midrule
Orig. &
\noindent Prior \textbf{information} and \textbf{generalized} questions: This paper ... uses a Bayesian decision theoretic \textbf{framework}, contrasting parallel and \textbf{inverse} decision problems, ... a subsequent \textbf{risk} minimization ... \\
\cmidrule(l{0pt}r{0pt}){1-2}
Adv.  &
\noindent Prior \textcolor{red}{knowledge} and \textcolor{red}{simplified} questions: This paper ... uses a Bayesian decision theoretic \textcolor{red}{system}, contrasting parallel and \textcolor{red}{opposite} decision problems, ... a subsequent \textcolor{red}{cost} minimization ... \\

\midrule[0.06em]

Orig. &
\noindent Several computer algorithms for discovering patterns in \textbf{groups} of protein sequences ... and these \textbf{algorithms} are sometimes \textbf{prone} to producing models that are \textbf{incorrect} because two or ... \\
\cmidrule(l{0pt}r{0pt}){1-2}
Adv.  &
\noindent Several computer algorithms for discovering patterns in \textcolor{red}{sets} of protein sequences ... and these \textcolor{red}{methods} are sometimes \textcolor{red}{vulnerable} to producing models that are \textcolor{red}{inaccurate} because two or ... \\

\bottomrule
\end{tabularx}
\end{table}

\section{Pseudo-code}
For completeness and clarity, the full procedures of the three IMDGA modules are provided in Algorithms~\ref{alg:topo_shap}–\ref{alg:edge_pruning} below.

\label{app:code}
\begin{algorithm}[!b]
\SetAlgoLined
\SetNoFillComment
\DontPrintSemicolon
\SetAlgoNoLine 
\SetAlgoNoEnd  
\SetKwInOut{Input}{Input}
\SetKwInOut{Output}{Output}
\SetKwInOut{Parameter}{Parameter}
\Input{Graph $\mathcal{G}=(\mathcal{V},\mathcal{E},\mathcal{X},\mathcal{T})$, target node $v\in\mathcal{V}$, victim Graph-LLM $\mathcal{F}_\theta(\cdot)$, text encoder $\psi_\theta(\cdot)$, tokenizer $Tokenize(\cdot)$, mask operation $Mask(\cdot)$, coalition sampler $\mathcal{S}(\cdot)$}
\Output{Pivotal word set $\mathcal{P}$}
\ForEach{$W_i \in \mathcal{W}$}{
    \ForEach{$S \in \mathcal{S}(\mathcal{W}\setminus\{W_i\})$}{
        $\mathcal{T}_{S\cup\{W_i\}}$ $\leftarrow$ $Mask(S\cup\{W_i\})$, \quad $\mathcal{T}_S$ $\leftarrow$ $Mask(S)$ \\ 
        $z_S \leftarrow \psi_\theta(\mathcal{T}_S)$, \quad $z_{S\cup\{W_i\}} \leftarrow \psi_\theta(\mathcal{T}_{S\cup\{W_i\}})$\\
        $f(\mathcal{T}_S)\leftarrow \sum_{u\in\{v\}\cup\mathcal{N}(v)} \mathcal{F}_\theta(\mathcal{G}, z_S, u)$ \\
        $f(\mathcal{T}_{S\cup i}) \leftarrow \sum_{u\in\{v\}\cup\mathcal{N}(v)} \mathcal{F}_\theta(\mathcal{G}, z_{S\cup\{W_i\}}, u)$ \\
        $w_{S} \leftarrow \dfrac{|S|!\,(m-|S|-1)!}{m!}$ \\
        $\phi(W_i) \leftarrow \phi(W_i) + w_{S}\cdot\bigl(f_S - f_{S\cup i}\bigr)$ \;
        \tcc{Compute SHAP contribution for word $W_i$}
    }
}
\ForEach{$W_i \in \mathcal{W}$}{
    $\xi(W_i) \leftarrow \sum_{u\in\{v\}\cup\mathcal{N}(v)} \phi^{y_u} (W_i)$ \;
    \tcc{Aggregate SHAP values over target node $v$ and its neighbors}
}

$\mathcal{I}_k \leftarrow$ indices of top-$k$ tokens by $\xi(\cdot)$ \;
\tcc{Select top-$k$ pivotal tokens }
$\mathcal{P} \leftarrow \{\,W_i \mid \xi(W_i) > \tau,\ i\in\mathcal{I}_k\,\}$ \;

\Return{$\mathcal{P}$}\;
\caption{Topological SHAP Module}
\label{alg:topo_shap}
\end{algorithm}

\begin{algorithm}[!b]
\SetAlgoLined
\SetNoFillComment
\DontPrintSemicolon
\SetAlgoNoLine 
\SetAlgoNoEnd  
\SetKwInOut{Input}{Input}
\SetKwInOut{Output}{Output}
\SetKwInOut{Parameter}{Parameter}

\Input{Graph $\mathcal{G}=(\mathcal{V},\mathcal{E},\mathcal{X},\mathcal{T})$, target node $v\in\mathcal{V}$, Graph-LLM $\mathcal{F}_\theta(\cdot)$, text encoder $\psi_\theta(\cdot)$, Masked Language Model (MLM), pivotal word set $\mathcal{P}$} 
\Output{Perturbed text $\mathcal{T}_v'$ for node $v$}
$\mathcal{C} \leftarrow \{\}$, \quad $count \leftarrow 0$ \;
\ForEach{$W_i \in \mathcal{P}$}{ 
    $\mathcal{C}[i] \leftarrow \text{top-}k$ candidates from MLM($W_i, \mathcal{T}_v$) \;
    \tcc{Generate top-$k$ replacement candidates for pivotal word $W_i$ using MLM}
}
\For {$i \leftarrow 0 $to $|\mathcal{P}|$ }{   
    \ForEach{$r \in \mathcal{C}[i]$}{
        $\mathcal{T}' \leftarrow (\mathcal{T}_v \setminus \{W_i\}) \cup \{r\}$ \;
        $p_u(\mathcal{G}, \mathcal{T}') \leftarrow \mathcal{F}_\theta(\mathcal{G}, \psi_\theta(\mathcal{T}'), u)$ \;
        $\delta_u(r) \leftarrow p_u^{(1)}(\mathcal{G}, \mathcal{T}') - p_u^{(2)}(\mathcal{G}, \mathcal{T}')$\;
        $\Delta(r) \leftarrow \sum_{u \in \{v\} \cup \mathcal{N}(v)} \delta_u(r)$\;
        \eIf{$\mathcal{F}_\theta(\mathcal{G}, \psi_\theta(\mathcal{T}'), v)$ flips label} {$\mathbb{I}_{\text{flip}}(r) \leftarrow 1$ }
        {$\mathbb{I}_{\text{flip}}(r) \leftarrow 0$}
        $\sigma(r) \leftarrow \Delta(r) \cdot (1 + \alpha \cdot \mathbb{I}_{\text{flip}}(r))$\;
    }
    \tcc{Compute score $\sigma(r)$ to select optimal replacement $r^*$}
    $r^* \leftarrow \arg\max_{r \in \mathcal{C}} \sigma(r)$\;
    $\mathcal{T}'_v \leftarrow (\mathcal{T}_v \setminus \{W_i\}) \cup \{r^*\}$\;
    $count  \leftarrow count +1$
    \lIf{$count > \beta \cdot|\mathcal{T}|$}{\Return{$T'_v$}}
    \tcc{Stop if modification exceeds ratio}
}

\Return{$\mathcal{T}'_v$}
\caption{Semantic Perturbation Module}
\label{alg:semantic_perturbation}
\end{algorithm}

\begin{algorithm}[hbp]
\SetAlgoLined
\SetNoFillComment
\DontPrintSemicolon
\SetAlgoNoLine 
\SetAlgoNoEnd  
\SetKwInOut{Input}{Input}
\SetKwInOut{Output}{Output}
\SetKwInOut{Parameter}{Parameter}

\Input{Graph $\mathcal{G}=(\mathcal{V},\mathcal{E},\mathcal{X},\mathcal{T}’)$, target node $v\in\mathcal{V}$, victim Graph-LLM $\mathcal{F}_\theta(\cdot)$, text encoder $\psi_\theta(\cdot)$, number of layers $k$, mask matrix $M \in \mathbb{R}^{k \times n}$, weight matrix $U$}
\Output{Pruned graph $\mathcal{G}'$}
\Parameter{weights $\alpha_1, \alpha_2, \alpha_3$}
$\mathcal{G}_n(v) \leftarrow \emptyset$
\ForEach{$u \in \mathcal{V}$}{
    $\delta(u) \leftarrow p_u^{(1)}(\mathcal{G}, \mathcal{T}_v') - p_u^{(2)}(\mathcal{G}, \mathcal{T}_v')$\;
    $I(u, v, k) \leftarrow \left\| \mathbb{E} \left[ (\partial \mathcal{X}_u^{(k)}) / (\partial \mathcal{X}_v^{(0)}) \right] \right\|_1$\;
    $I_u(v, k) \leftarrow \frac{I(u, v, k)}{\sum_{w \in \mathcal{V}} I(u, w, k)}$\;
    $\text{Score}(u) \leftarrow \alpha_1 \cdot (1 - \delta(u)) + \alpha_2 \cdot I_u(v, k) + \alpha_3 \cdot \left( \frac{1}{\deg(u)} \right)$}
    \tcc{Calculate the weighted score of predictive disparity, feature, and degree.}
$\mathcal{G}_n(v) \leftarrow$ top-$k$ nodes by $\text{Score}(u)$\;
$\hat{\phi} \leftarrow (M^\top U M)^{-1} M^\top U \hat{y}$\;  
\ForEach{edge $e \in \mathcal{E}$ in $\mathcal{G}_n(v)$}{
    Assign attribution $\phi_e$ from $\hat{\phi}$}
$\mathcal{E}' \leftarrow \mathcal{E} \setminus$ top-$k$ edges by $\phi_e$\; 
$\mathcal{G}' \leftarrow (\mathcal{V}, \mathcal{E}', \mathcal{X}, \mathcal{T})$

\Return{$\mathcal{G}'$}
\caption{Edge Pruning Module}
\label{alg:edge_pruning}
\end{algorithm}

\end{document}